\pdfoutput=1

\documentclass[11pt]{article}

\usepackage{ACL}

\usepackage{times}
\usepackage{latexsym}

\usepackage[T1]{fontenc}

\usepackage[utf8]{inputenc}

\usepackage{microtype}

\usepackage{inconsolata}

\usepackage{graphicx}

\usepackage{natbib}

\usepackage{xspace}
\usepackage{booktabs}
\usepackage{multicol}
\usepackage{amsmath}
\usepackage{multirow}
\usepackage{graphicx}
\usepackage{enumitem}
\usepackage{mdwlist}
\usepackage{textcomp}
\usepackage{siunitx}
\usepackage{longtable}
\usepackage{verbatim}

\usepackage{caption}
\usepackage{subcaption}

\usepackage{multicol}
\usepackage{longtable}

\usepackage{csvsimple}
\usepackage{longtable}

\newcommand{\acronym}{\textsc{SPoRC}\xspace}

\newcommand{\sref}[1]{\S\ref{#1}}

\newcolumntype{L}[1]{>{\raggedright\let\newline\\\arraybackslash\hspace{0pt}}m{#1}}
\newcolumntype{C}[1]{>{\centering\let\newline\\\arraybackslash\hspace{0pt}}m{#1}}
\newcolumntype{R}[1]{>{\raggedleft\let\newline\\\arraybackslash\hspace{0pt}}m{#1}}

\title{Mapping the Podcast Ecosystem with the\\Structured Podcast Research Corpus}

\author{Benjamin Litterer\\
    School of Information \\
    University of Michigan \\
  \texttt{blitt@umich.edu} 
  \\\And
  David Jurgens \\
    School of Information and \\ Computer Science \& Engineering \\
    University of Michigan \\
  \texttt{jurgens@umich.edu} 
  \\\And
  Dallas Card \\
  School of Information \\
    University of Michigan \\
  \texttt{dalc@umich.edu} 
  }

\begin{document}

\maketitle

\begin{abstract}
Podcasts provide highly diverse content to a massive listener base through a unique on-demand modality. However, limited data has prevented large-scale computational analysis of the podcast ecosystem. To fill this gap, we introduce a massive dataset of over 1.1M podcast transcripts that is largely comprehensive of all English language podcasts available through public RSS feeds from May and June of 2020. This data is not limited to text, but includes metadata, inferred speaker roles, and audio features and speaker turns for a subset of 370K episodes.
Using this data, we conduct a foundational investigation into the content, structure, and responsiveness of this ecosystem. Together, our data and analyses open the door to continued computational research of this popular and impactful medium.

\end{abstract}

\section{Introduction}
\label{sec:introduction}

Over the past few decades, podcasts have emerged as an important part of the modern media landscape \citep{pewPodcastsNewsInformation}. Despite their growing importance, most published research on podcasts has used small samples and/or non-computational methods to study topics like educational content \citep{quintana.2021.podcasts}, misinformation \citep{brookingsAudibleReckoning}, and listener preferences \citep{moe.2022.podvertising}, while largely neglecting distinguishing aspects of this medium, such as its audio characteristics and social network.

To enable research into media, communication, and the podcast ecosystem, we introduce the Structured Podcast Research Corpus (\acronym), the first large-scale open dataset of podcasts providing a nearly comprehensive slice of publicly available data to support computational social science research. In this paper, we present \acronym and our process for building it, along with preliminary analyses of the dataset, exploring the breadth of topical content, network linkage across shows, and responsiveness to real-world news events.

Given that podcasts are already a key channel for people to consume entertainment and information \citep{tobin.2022.why,pewPodcastsFactSheet}, one cannot fully understand the modern media landscape without considering long-form audio content. 
However, in comparison to domains like news or social media, there is currently a dearth of available datasets for studying this medium at scale. %
By creating a large-scale corpus that surfaces both text and audio aspects of podcasts, enhanced with additional metadata, we aim to facilitate the kinds of insights that have come from similarly scaled public research datasets for Twitter \citep{pfeffer2023just}, Reddit \citep{baumgartner2020pushshiftredditdataset}, Amazon \citep{ni-etal-2019-justifying}, and other domains.

While the long term evolution of podcasts is also of great interest,
the initial \acronym dataset is a thick slice of shows, representing every English language episode we were able to retrieve from the months of May and June 2020---a total of 1.1M episodes with transcripts and basic metadata. Diarization and speaker identification have also been included for a subset of episodes.
This sample thus provides a foundation for future understanding, including the long tail of podcast content.

In the rest of the paper, we first introduce \acronym and the methods used to create it (\S\ref{sec:dataset}).
We then characterize this snapshot of the podcast ecosystem in terms of content and structure, revealing the distribution of topics discussed on podcasts, and the community network structure created by guest co-appearances (\S\ref{sec:ecosystem}).
Finally, we analyze the responsiveness of podcasts, examining the timing and extent of the impact of a major media event (\S\ref{sec:georgeFloyd}), finding varied responses across different segments of the medium.
Overall, our results point to new questions for further research into community identity, information diffusion, and incidental news exposure in podcasts, which our release of \acronym will enable. Our data and code are made available for non-commercial use and can be found at \href{https://github.com/blitt2018/SPoRC_data}{https://github.com/blitt2018/SPoRC\_data} and \href{https://huggingface.co/datasets/blitt/SPoRC}{https://huggingface.co/datasets/blitt/SPoRC}.

\section{Background}
\label{sec:background}

Podcasts are a massively popular, diverse, and impactful medium. In 2023, 42\% of Americans aged 12 and older reported having listened to a podcast in the past month, with many listening multiple times per week \cite{pewPodcastsFactSheet}. %
Podcasts feature long-form educational content alongside short morning news shows, true crime dramas, life advice, and comedy. These shows also have real-world impact; listeners report shifting their media diet (60\%), changing their lifestyle (36\%), buying a product (28\%), or contributing to a political cause (13\%), as a result of their consumption \cite{pewPodcastsNewsInformation}. 
Importantly, the majority of podcast listeners trust information they hear on these shows \cite{pewPodcastsNewsInformation}, even as evidence accumulates about podcasts spreading misinformation \citep{brandt.2023.echoes,brookingsTheChallenge,brookingsAudibleReckoning}.

Podcasts are also interesting due to the low startup costs and accessibility of the medium. %
Although led by independents, the medium was quickly adopted by legacy media institutions, who had huge success with shows like \emph{Serial} \cite{berry2006, markman2012, markman2014,berry_golden_2015}. 
Although broad in coverage, various studies have pointed to a lack of diverse voices in podcasting, with, for example, men being overrepresented among both overall listeners and the hosts of top shows \citep{werner.2020.women}.

Despite their popularity and reach, there has been surprisingly little large-scale academic research into podcasts, in part due to a lack of data. The largest existing research corpus is a dataset of 200K episodes (half in English; half in Portuguese) released by Spotify \cite{spotify}. But that dataset is no longer being maintained, and Spotify is no longer granting access to new users. A smaller corpus, with 240 hours of podcasts, annotated with emotion labels, was released in 2019 \citep{lotfian.2019.building}.
Another project, PodcastRE, is actively preserving podcast audio, but the focus of that project is primarily archival, and it does not widely distribute data for research \cite{morris.2019.podcastre}. Finally, the Brookings Institute has assembled and made available the Popular Political Podcasts Dataset (PPPD), a corpus of around 120 prominent political podcast, which is actively being maintained \citep{brandt.2023.echoes}.

As with comparable resources for other types of data sources, such as Congressional Speeches \citep{gentzkow.2019.measuring} or Reddit \citep{baumgartner2020pushshiftredditdataset}, 
our dataset enables new research into podcasts, with speaker-labeled content over a subset that is dense enough for the study of communities, yet long enough to study temporal trends. 
Compared to other media, our dataset captures long-form conversations, augmented with audio features that give insight into how each speaker is communicating, with speakers appearing in many venues, connected via content, guests, and other mechanisms.
Towards this end, we carry out two initial high-level analyses (\S\ref{sec:ecosystem} and \S\ref{sec:georgeFloyd}) that make use of multiple dimensions of this data to study the structure and responsiveness of the podcast ecosystem. We also intend that the pipeline we have created (\S\ref{sec:dataset}) will be useful to others in curating additional podcast datasets.

\section{Building a Massive Podcast Dataset}
\label{sec:dataset}

To create a comprehensive and open dataset for research into the podcast ecosystem, or for use in studying related social scientific and linguistic questions, we set out to collect and process all extant podcast content available from RSS feeds, covering a fixed slice of time.\footnote{For the purposes of this work, we consider podcasts to be any serially-released audio recordings distributed as mp3 files via RSS feeds; note that this excludes video podcasts that are \emph{only} distributed on platforms such as YouTube. Here, we use the term ``podcast'' and ``show'' synonymously, where each podcast consists of one or more episodes. For additional discussion of how to define the medium, see \citet{mcgregor.2022.podcast}.}\textsuperscript{,}\footnote{RSS feeds are the standardized format used to structure podcast metadata, allowing them to be streamed across a variety of platforms. Although the vast majority of podcasts are freely available and portable in this way, a subset have exclusive deals with specific hosting platforms preventing public access to their feeds.} 
Although long term dynamics of podcasts are also of great interest, \acronym provides a foundation for future exploration, and can easily be extended by others using code that we will release.
Because our goal is not to redistribute audio files, we only preserve and share transformed versions of original podcast audio, as described below.

\subsection{Initial Data Collection}

To bound the scope of the initial dataset, we chose the months of May and June, 2020, and attempt to collect all podcast audio files from that time that are still available from RSS feeds.\footnote{Choosing a time period from the recent past allows us to understand the data with some context; notable events during this time period include the murder of George Floyd and the ensuing protests, the start of Operation Warp Speed, and US deaths from COVID-19 passing 100K people.}
To identify RSS feeds, we start with data from 
Podcast Index, a public database which includes information on over 4 million different shows, complete with podcast-level metadata.\footnote{\url{https://podcastindex.org/}}
Using the RSS feeds provided by Podcast Index, we identify  273K English-language shows that released episodes during May or June, 2020.
From these feeds, we are able to download both audio files and episode-level metadata for a total of 1.3M episodes from 247K different shows.

\subsection{Transcription}
To extract the words that are spoken, we begin by transcribing each audio file to a plain text format.
For this, we use Whisper, a publicly available automated speech recognition (ASR) system \citep{radford2022robustspeechrecognitionlargescale}. Specifically, we use the \texttt{whisper-base.en} model, which represents a compromise between quality and speed.\footnote{\href{https://huggingface.co/openai/whisper-base.en}{\texttt{https://huggingface.co/openai/whisper-base.en}}} Whisper also provides approximate timestamps at the word-level, as well as tags such as ``(laughing)'' and ``[MUSIC]'' to mark non-speech audio segments. 

Although prior work has identified some important issues with Whisper, including a propensity to hallucinate continuations \citep{koenecke.2024.careless}, we find that this is not a major issue for our data. Rather, the biggest source of errors is that the model will sometimes repeat short phrases many times in a row during segments featuring music, silence, or non-English speech. To maintain high quality transcripts, we use a simple n-gram filter to identify and remove episodes that have a high proportion of this sort of repetition. Regardless, overall transcription quality remains high: validation against a sample of professional podcast transcripts reveals word error rates of less than 10\% on average. Moreover, closer inspection reveals that more than half of these errors are due to the way the professional transcripts have been edited for fluency. (See Appendix \ref{app:transcription} for details).  After filter, we are left with 1.1 million episode transcripts, comprising 6.6 billion words. 

\subsection{Prosodic Feature Extraction}
In addition to textual content, podcast audio contains rich information about how a speaker is communicating. To obtain this additional information, we use the openSMILE toolkit \citep{EybenOpenSMILE}, which is widely used for audio feature extraction in scientific publications.\footnote{\url{https://www.audeering.com/research/opensmile/}}
We focus on a subset of the eGEMAPS feature set \citep{eyben2015geneva}; in particular, we extract the fundamental frequency (F0), the first formant (F1), and the first four Mel Frequency Cepstral Coefficients (MFCCs 1-4).
Each features was chosen for its relevance to sociolinguistic phenomena. 

\textbf{F0} is related to vocal pitch, and captures the lowest frequency of oscillations of a voiced sound. In addition to varying across speakers, speakers manipulate this frequency in speaking, such as to mark emphasis or questions. 
Formants capture additional resonances or harmonics, with \textbf{F1} being commonly associated with vowel pronunciation. It has been further employed to describe variation in speakers' dialect, sex, and age \cite{hagiwara1997, kent2018}. Finally, \textbf{MFCCs} are features designed to capture the short-term power spectrum of sounds, and can be used in concert with other features to measure general vocal characteristics for a number of downstream tasks such as emotion recognition \cite{luengo2005}. openSMILE measures these features at a high frequency which incurs a large storage requirement; therefore, we collapse these to the token-level, keeping the mean of each feature over each word's duration.

\subsection{Identifying Speakers}
\label{sec:identifying-speakers}

A key limitation of Whisper is that it does not distinguish between speakers. In order to measure turn-level information from podcast conversations, we attempt to match speaking turns to speakers, using a combination of off-the-shelf and newly developed tools.

\paragraph{Speaker Diarization:}
As a first step, we use \texttt{pyannote} \citep{Bredin23}  to split audio files into individual speaker turns, using a process known as diarization.\footnote{Due to computational limitations and the speed of diarization, we only diarize 370K episodes in our initial dataset release, sampled randomly from the full set of transcribed episodes, but will update this in subsequent data releases.}
\texttt{pyannote} determines the number of distinct speakers in a conversation, matching these speakers to generic labels (e.g., speaker$_1$, speaker$_2$, etc.).
Each audio segment is then assigned to one of these speakers, which allows us to map each token from transcripts to a corresponding speaker turn, using audio timing information.
Validation against a sample of professional transcripts reveals a diarization word error rate of 2.1\% on average for a small sample of two-person conversations.
(See Appendix \ref{app:diarization} for details).
After excluding speakers that make up less than 5\% of the total speaking time in an episode, we find that 37\% of episodes in our diarized data have only a single speaker, with 39\% having two speakers, and the rest having $\ge$3. 

\paragraph{Identifying Host and Guest Names}
Unfortunately, mapping generic speaker IDs to named speakers or speaker roles is difficult, as this information is not reliably included in episode metadata in a consistent format. 
As a step towards this, we develop a model to identify the names and voices of the host(s) and guest(s) for each episode. 

After preliminary exploration, we determined that transcripts were the most reliable sources of this information, as hosts and guests are typically introduced by name (e.g., ``I am your host ...'', etc.).\footnote{In addition, some hosts never introduce themselves, or provide only their first name, and thus remain effectively anonymous in our data.}
To determine speaker names and match them to voices, we develop a pipeline to first identify  candidate names in text, and then classify each of these names as \textsc{Host}, \textsc{Guest}, or \textsc{Neither}, with the third category representing people mentioned but not appearing in the episode.
To make these assignments, we rely on the direct and indirect linguistic cues surrounding mentions of a speaker's name that indicate their role. We first use spaCy \cite{honnibal2020spacy} to identify all named \textsc{Person} entities that occur in an episode description, or in the first 350 words of the transcript, filtering this list to keep only mentions that include a first and last name. %

To generate training data for this task, we collect three human judgments from Profilic annotators on a subset of 2,000 candidate entities in context. Annotators label one target entity at a time, see the  podcast description, episode description, and 300 words of the transcript, and provide a judgment of \textsc{Host}, \textsc{Guest}, or \textsc{Neither}.
Despite some ambiguous cases, annotators had relatively high chance-corrected agreement on this task (Krippendorff's $\alpha$=0.77). 
Final labels were obtained by aggregating annotations using MACE \cite{hovy2013learning}, resulting in labels for 858 \textsc{Hosts}, 639 \textsc{Guests}, and 503 \textsc{Neither}. 
Using these annotations, we fine tune a RoBERTa model to classify a named entity based on its mean-pooled embeddings, which achieves 0.87 cross-validation accuracy and 0.88 accuracy on a held out test set. 
We then apply this model to all episodes for which there is at least one identified candidate entity.
For diarized podcasts where our model identifies exactly one of the speakers as the \textsc{Host}, we heuristically map that person to the first voice which says the host's name.
For additional details on speaker labeling, see Appendix \ref{app:role-labeling}.

\paragraph{Results} Using our role annotation model, we identify participants in all episodes; over 550K episodes (49\%) of the dataset have at least one identified \textsc{Host} or \textsc{Guest}, including 386K \textsc{Host} names and 535K \textsc{Guest} names. Based on these inferred participants, we find that the plurality of podcasts have a single \textsc{Host}, followed by a single \textsc{Host} and single \textsc{Guest}. (See Figure \ref{fig:hostCount} in the Appendix). During manual review, we found a few notable false positives, such as Ryan Reynolds, who appears in a common podcast ad in which he is labeled as a \textsc{Host}. Nevertheless, we verify that some guests truly are prolific, such as Matt Ridley, who appears as a guest on 21 episodes in our data.
Further refinements could improve on our host and guest identification by combining information across episodes and using external knowledge bases, but these labels nevertheless permit further investigation into how guests connect shows across categories (\S\ref{sec:ecosystem}), and we leave such extensions of our model as future work. %

\subsection{Summary of Our Dataset}
Altogether, our data includes rich multimodal information spanning the episode and speaker-turn levels. At the episode level, we provide transcript text alongside category, duration,  publication date, and inferred host and guest information. At the turn level, we provide transcripts split into segments of speech corresponding to different speakers, start and end timestamps, averaged audio features for each segment, and their inferred host and guest roles where possible. We release episode-level data for the full set of 1.1 million podcast episodes and speaker turn data for the subset of 370K diarized episodes (see Appendix \ref{descriptiveOverview} for details).

\section{Mapping The Podcast Ecosystem}
\label{sec:ecosystem}

Although there are many studies of podcast content \cite{lindgren2016narrative,drew2017Educational,little2020medicine,morris.2021.saving}, advertising \cite{moe.2022.podvertising,brinson2023investigating, bezbaruah2023podcast}, audiences \cite{whipple2023examining, werner.2020.women}, and impacts \cite{brookingsAudibleReckoning, pewPodcastsNewsInformation}, all of these have been carried out at a small scale, often relying on curated samples and/or manual annotation. To provide the first comprehensive documentation of the overall podcast ecosystem, we begin by characterizing its content and network structure. 
In particular, we focus on two questions: 1) What are the major content areas discussed on podcasts? and 2) To what extent are communities isolated or connected across different categories in terms of their topical coverage and common guests? Here, we distinguish between topical and networked communities, where \emph{topical communities} are podcasts with similar content, and \emph{networked communities} are podcasts linked through common guests.

To answer these questions, we consider both podcasts' content and the social network formed by common guests. 
Our topical approach is motivated by the fact that information in podcasts has had both helpful and pernicious effects on society. In the medical domain, for example, podcasts have served as a tool for education and support \cite{billman2024developing, hurst2019podcasting} as well as source of dangerous misinformation, particularly during the height of the COVID-19 pandemic \cite{brookingsAudibleReckoning}. Our analysis here helps to reveal, for the first time at this scale, the overall prevalance of various topics of discussion on podcasts, both within and across categories.

To complement our topical analysis, we capture the social network of this ecosystem, using guests as key links between shows. Prior work has found that edges in social networks facilitate the diffusion of information \cite{Bakshy2012role}. Furthermore, guest invitations reflect the preferences of podcast hosts and their audiences. Thus, podcasts linked through shared guests may play a role in forming networked communities with a shared identity or set of common knowledge \citep{pewGuestsReport}. In contemporaneous work, \citet{demets.2025.podcasts} have also looked at both the guest co-occurrence network, and the movement of guests between shows, using the PPPD corpus of around 120 mostly political podcasts.
Whereas topics are the sole focus of our first research question, we use both topical information and this social network to answer the second.

\paragraph{Methods}
Podcasts have self-selected high-level category labels (e.g., \textsc{News}, \textsc{Sports}, \textsc{Society}), with most shows using well established primary categories. To understand the interaction between content and category, we augment these labels to directly model the content in episodes. We fit an LDA topic model on the first 1000 words of each podcast episode transcript with 200 topics, using Mallet \citep{blei.2003.latent,mccallum.2002.mallet}. Each episode is then represented as its topic distribution, allowing us to compare episodes in terms of topical similarity.

While topics provide one useful means for finding related podcasts, a distinct type of connection is indicated by the appearance of the same guest on multiple different shows. Here, the podcast guest network can be formed by first constructing a bipartite graph $(P, G, E)$ with an edge $e_i \in E$ connecting a podcast $p_j \in P$ to a guest $g_k \in G$ when that guest appears on any episode of the podcast.

Guests are identified as any two-word named entity labeled as \textsc{Guest} by our model, as described in \S\ref{sec:identifying-speakers}. To minimize the effect of false positives, we exclude guests that have names among the top 50\% of most frequent named entities (e.g., John Smith) as well as all names classified as a \textsc{Host} in a different episode of the same podcast. We then project this bipartite network to a one-mode network where podcasts are connected if they share one or more guests in common, which we refer to as the podcast-guest network. The resulting network has 10,480 vertices (podcasts) and 26,589 edges (connections based on co-appearing guests). For details, see Appendix \ref{app:network}.

\begin{figure*}[h]
\centering
\includegraphics[width=\textwidth]{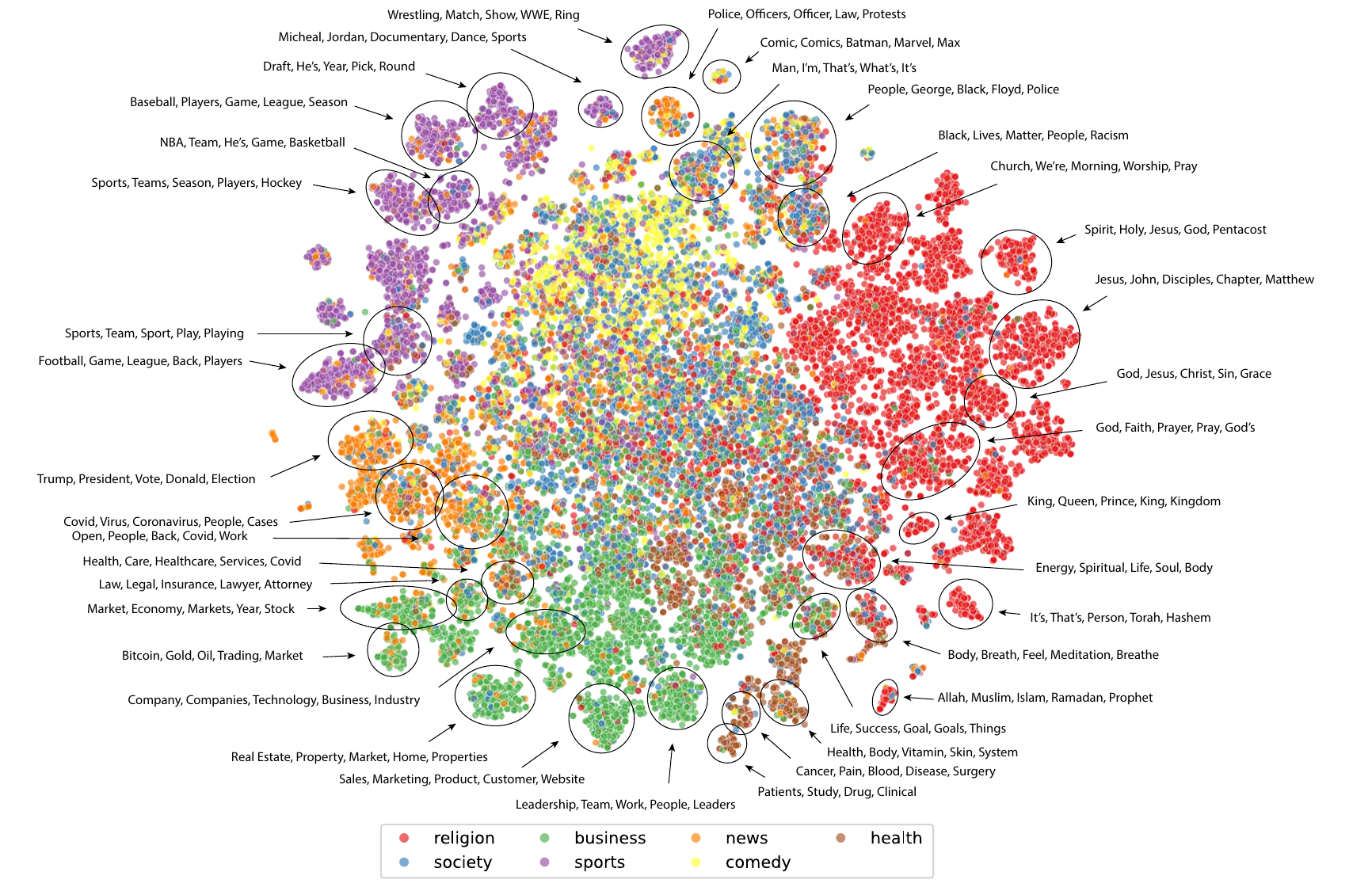}
\caption{Many topics are strongly associated with a single category. However, a number of topics such as ``Black, Lives, Matter'', and ``Life, Success, Goals'' cut across categories. Here, this is depicted using a sample of 25K episodes, colored by category, and projected using t-SNE on episodes' topic distributions to visualize topical distance, with select topic clusters annotated using the top words in the corresponding topic.} 
\label{topicFig}
\end{figure*}

\paragraph{Topical Communities}
To address the topical dimension of our research questions, we visualize podcasts with respect to their category label and content similarity, revealing both coherent topical communities as well as some that span multiple categories.
Many topical communities fall neatly within their self-ascribed category labels, indicating that these categories are a meaningful descriptor of the type of content within podcasts. This is particularly true for categories such as \textsc{Sports}, \textsc{Religion}, and \textsc{Business}, each of which have coherent thematic sub-communities such as Baseball, Wrestling, Real Estate, Bitcoin, Judaism, and Islam.\footnote{Surprisingly, \textsc{Religion} turns out to be the most common podcast category in our dataset, with many of these shows consisting of recorded Christian sermons.} Other broad topics appear to be further fragmented, such as Christianity, for which a number of topics appear to correspond to different parts of the liturgy or themes within Christianity. 

By contrast, however, there are a number of topical communities that include creators from multiple categories or supersede category boundaries entirely. Two such communities focus on spirituality and self-improvement, which include content from episodes in the \textsc{Religion}, \textsc{Society}, and \textsc{Business} categories and the \textsc{Business}, \textsc{Religion}, and \textsc{Health} categories respectively. This is also the case for topics related to COVID-19, as well as racial justice, which is explored in-depth in \S\ref{sec:georgeFloyd}. The presence of such topical communities suggests that existing category labels do not fully describe the meaningful groupings in podcast content, and that these are the areas in which we might expect to see the most extensive cross-fertilization and exchange of ideas.

\paragraph{The Structure of the Podcast Social Network}
Guests play an important role in structuring parts of the podcast ecosystem. 
High-profile guests may appear frequently on multiple podcasts to promote their brand or ideology \cite{bratcher2024codeSwitching}, facilitating the exchange or even diffusion of ideas. 
As a result, guest appearances are often a strong indicator of what type of content hosts prefer to feature and what will appeal to their audience.

\begin{figure}[t!]
\centering
\includegraphics[width=\columnwidth]{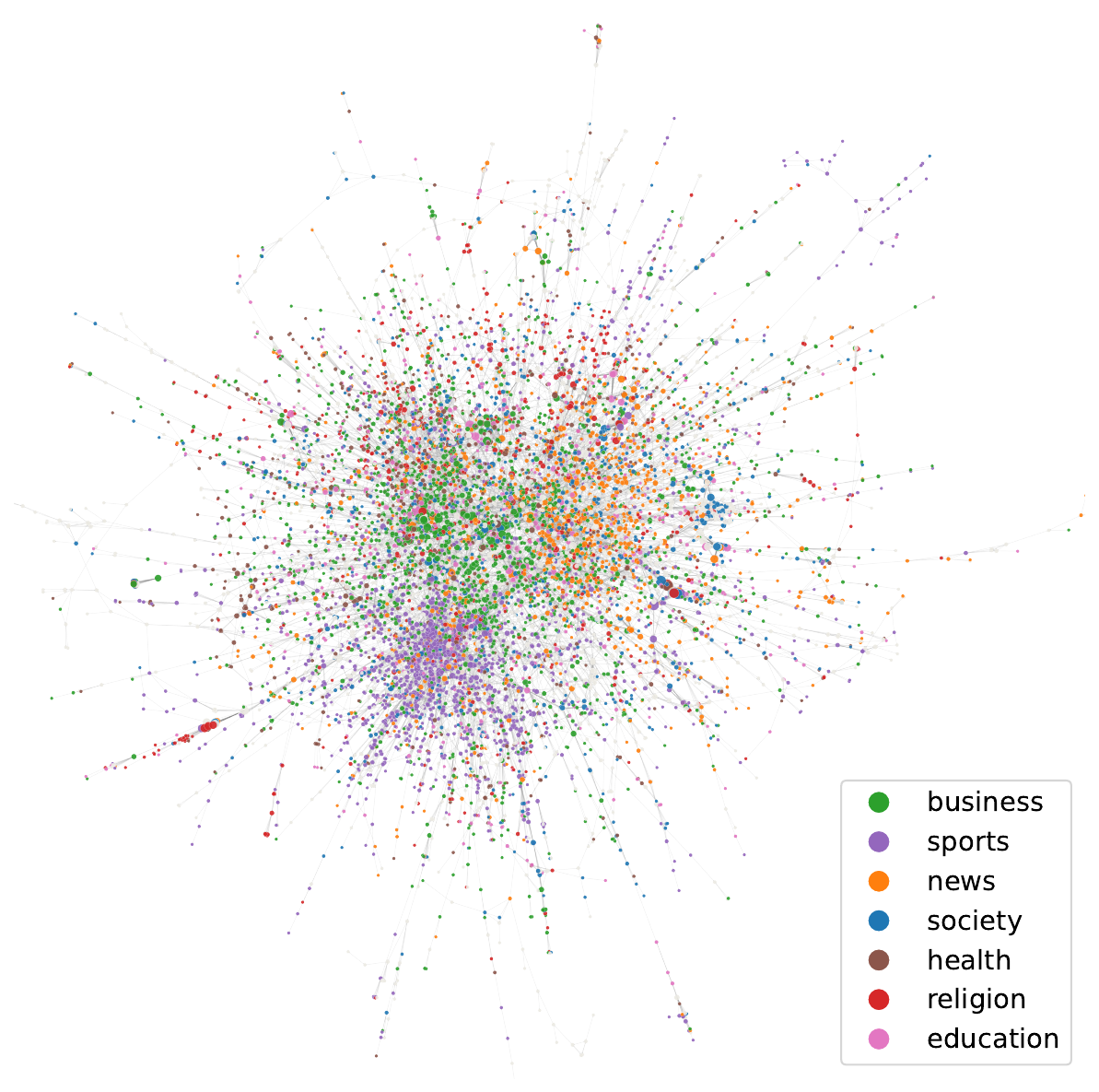}
\caption{\textsc{Business}, \textsc{Sports}, and \textsc{News} have densely connected guest networks, with other categories being more diffuse.
Edges in this network connect podcasts that share one or more common guests. Nodes represent podcasts, with color mapped to category, and node size indicating a podcast's total number of shared guests.}
\label{fig:guest_network}
\end{figure}

We find that the association between guest appearances and category label varies significantly across categories, as shown by the podcast-guest network in Figure \ref{fig:guest_network}. Here, we visualize the network with nodes colored by category, positioned by the force-directed Yifan Hu algorithm, and sized proportionate to their edge degree \cite{yifan2005}. The figure shows that some categories such as \textsc{Business}, \textsc{News}, and \textsc{Sports} correspond to relatively distinct networked communities---i.e., podcasts within the same category share a distinct pool of guests that do not interact much with podcasts from different categories.

Despite being the largest categories, \textsc{Religion} and \textsc{Society} do not form large networked-communities in our network. As shown in supplemental Figure \ref{fig:guestCount}, both the \textsc{Religion} and \textsc{Society} categories appear to invite guests substantially less frequently than other large podcast categories, and thus provide fewer opportunities for cross-fertilization of ideas. This suggests that creators may be less well connected for these categories, though a longitudinal sample may help identify such structure.

\begin{figure*}[t!]
\centering
\includegraphics[width=0.93\textwidth]{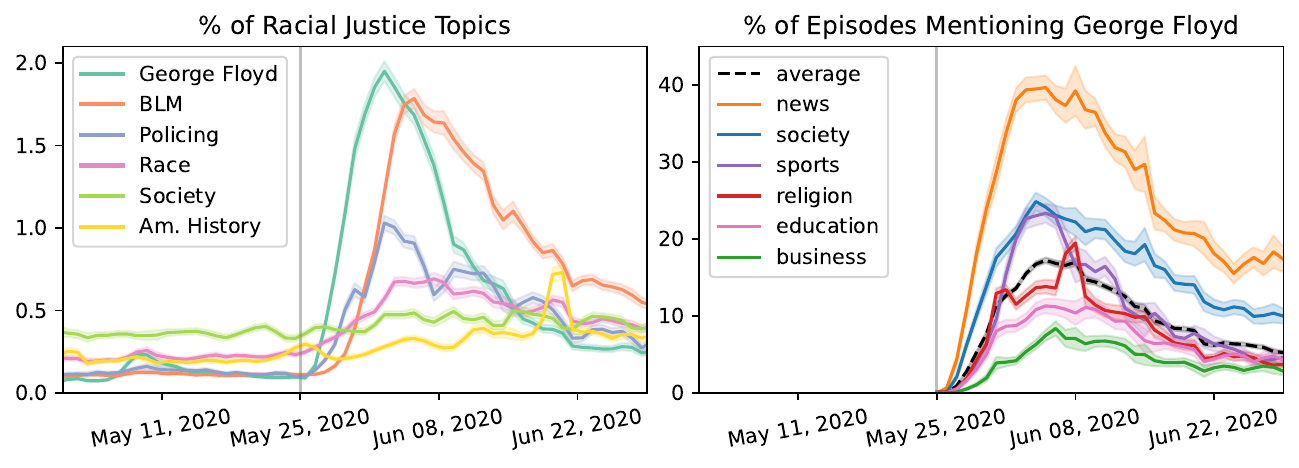}
\caption{The murder of George Floyd triggers a fast and widespread discussion of racial justice in the podcast ecosystem. On the left, we plot a three day rolling average of the topic percentages across all transcripts. On the right, we plot a three day rolling average over the percentage of episodes where the name George Floyd was said. Shaded bands represent 95\% confidence intervals.}
\label{fig:2panelFloydSummary}
\end{figure*}

\begin{table}[t!]
\centering
\small
\begin{tabular}{ rcc } %
\textbf{Category} & \textbf{Modularity} \\ \hline
\textsc{Sports}  & 0.155   \\ %
\textsc{Business}  & 0.134   \\ %
\textsc{News}   & 0.064   \\ %
\textsc{Religion} & 0.045   \\ %
\textsc{Society}  & 0.013   \\ %
\textsc{Education} & 0.011  \\ %
\end{tabular}
\caption{\textsc{Sports} and \textsc{Business} form particularly dense modules on the basis of common guests, as demonstrated by their high modularity. This trend indicates that these categories are more insular, and substantially more likely to form connections within themselves than would be expected by random chance.}
\label{tab:modularity}
\end{table}

To quantify the relationship between category labels and identifiable networked communities, we calculate modularity for the partitions of our network corresponding to different categories to measure their intraconnectedness, as reported Table \ref{tab:modularity}. This measures the extent to which membership in a category increases the probability of two nodes sharing a common guest relative to a graph with randomized edges (see Appendix \ref{app:modularity} for details).
The high modularity of \textsc{Business} and \textsc{Sports} suggests that podcasts within these categories are particularly likely to share guests relative to podcasts in other categories.

\paragraph{Summary} Overall, we find evidence that categories provide meaningful structure to the podcast ecosystem. With respect to content, we find many distinct fandoms represented (e.g., Wrestling, Bitcoin), with many such topical communities belonging to a single category. However, a number of high-stakes topics (i.e. COVID-19, racial justice, and wellness) have cross-category discussion, a potential indicator that these domains may serve as a ground for discussion, debate, or the diffusion of ideas. With respect to guests, we find that \textsc{Business} and \textsc{Sports} form networked communities, whereas two of our largest categories (\textsc{Religion}, \textsc{Society}) do not. Our results cannot by themselves inform how community structure influences discourse in the podcast sphere. However, they answer the most fundamental question for future work by telling us where this community structure exists.

\section{Collective Attention in the Ecosystem}
\label{sec:georgeFloyd}

In news domains, media environments often respond rapidly to major events, setting the agenda for public discussion and debate \cite{king_how_2017, boydstun_two_2014}. Due to their ubiquity, podcasts may play an equally important role in agenda-setting. However, given its fragmentation, it's not currently known how responsive the ecosystem is to current events. Here, we test for collective attention using a case study on George Floyd's murder, measuring both the (1) temporal responsiveness and (2) overall diffusion associated with this event. George Floyd's death was one of the most significant events of that year in terms of news coverage \cite{cowart2022framing,reny2021opinion}, and, as it occurs in the middle of the \acronym time frame, provides an ideal test case of the response of the podcast ecosystem to real-world events. 

\paragraph{Methods}
To identify episodes relating to the case study's themes, we manually identify topics from our previous analysis (\sref{sec:ecosystem}) which relate either to the murder of George Floyd or racial justice more broadly. Six such topics were identified, corresponding roughly to Black Lives Matter, George Floyd, Race, Society, Policing, and American History. We measure the mean topic distribution over time for podcasts in each category to assess whether the ecosystem has increased its collective attention. 
Discussion coverage is measured as the daily percent of episodes with the name ``George Floyd'' in their transcript.

\paragraph{Results}
Collective attention in podcasts responds rather quickly to the initial event, then decreases far slower over the following weeks (Figure \ref{fig:2panelFloydSummary}). In particular, the percentages for both the George Floyd and Black Lives Matter topics rises to their peaks over the course of roughly 10 days, with a lag of approximately 4 days in-between them. This pattern of quick story development and slow story decay aligns with the phenomenon of a ``media storm'' in the news medium \cite{boydstun_two_2014}. However, this case suggests that development of media storms in podcasts may operate at a slower pace, as prior computational work on news media found that the average storm had peaked by day three and decayed by day ten \cite{litterer2023rainspours}. Given that there are substantially more podcasts than media outlets (in part due to ease of entry), future work is needed to understand how the two ecosystems differ in their responses.

This event was widely discussed, with 21\% of shows saying the name ``George Floyd'' in one or more episodes up to the end of June. Notably, these mentions were not limited to categories that typically discuss current events or politics (e.g., \textsc{news} or \textsc{society}); rather, all categories have significant discussion of George Floyd at their peaks, but only categories which would routinely discuss these issues have elevated mentions at the end of our time period. This trend is seen clearly in \textsc{society} and \textsc{sports}. While both categories had $\sim$20\% of their episodes mentioning Floyd at their peak, \textsc{society}'s mention percentage was over twice that of \textsc{sports} at the end of our time-period. 

To further investigate how podcast categories differ in their response to George Floyd's murder, we consider the time series of topics within each category (Figure \ref{fig:categoryTimeSeries}). We find that some categories differ in the relative emphasis placed on different topics. Most notably, \textsc{news} is the only category to give significant attention to the topic concerning policing, law, and protest. This finding aligns with other work on discussions of race in the Summer of 2020, which found that discussion of law and order was primarily driven by white journalists \cite{le2022blm}. 

Our analyses confirm that 1) the podcast ecosystem's response had similar temporal patterns to news media, and that 2) discussion of George Floyd was widespread, yet variable, across communities. In the first case, our finding builds off of work on ``media storms'' and builds toward future work into the underlying mechanisms through which collective attention in podcasts rapidly shift and slowly decay. With regard to the widespread discussion of George Floyd, our work is broadly supported by findings that political discussion is abundant across communities with both political and non-political labels \cite{rajadesingan_political_2021, munson2011prevalence}. This suggests that, as with other platforms, listeners may be incidentally exposed to political content from hosts whom they trust and feel connected to \cite{pewPodcastsNewsInformation,fletcher2018incidental, weeks2020ecology}.

\section{Discussion}
\label{sec:discussion}

While it is the largest analysis of its kind, ours is not the first work to computationally analyze podcast data. Existing computational work on podcasts has touched on aspects like summarization \citep{vaiani.2022.leveraging, vartakavi2021}, identifying misinformation \cite{cherumanal2024everything}, predicting popularity \citep{joshi-etal-2020-read}, and identifying narratives \citep{abdessamed-etal-2024-identifying}. Much of this research has used the Spotify dataset \cite{spotify} and has been formulated in terms of specific NLP tasks. While being applicable to social science questions as well, our data has similar applications to NLP but is larger and captures more features.  

In addition to NLP applications, our data and analyses enable deeper investigation into social science questions. Section \ref{sec:ecosystem} emphasized structure, identifying communities within podcasts. Future research could investigate the political and social characteristics of these communities, similarly to past work on the Reddit and News ecosystems \cite{waller2021RedditCommunities, niculae2015quotus}. For example, our textual and prosodic information could be used to infer whether communities form along the lines of gender performance and identity in this male-dominated space. 

In section \ref{sec:georgeFloyd}, we illustrate collective attention dynamics in the podcast ecosystem through our study of racial justice discussions in 2020. Rather than focusing on a single event, future work may consider how this ecosystem responds to a number of politically salient topics, extending work on news media to the podcast ecosystem \cite{boydstun_two_2014, litterer2023rainspours}. We found that discussion of George Floyd permeated even categories such as \textsc{religion} and \textsc{business}. Building on this finding, future work could model how information diffuses across communities and under what conditions this is likely to occur.

\section{Conclusion}

Podcasts are a widely distributed, unique, and impactful form of media, but comprehensive data for the computational study of this medium have been lacking.  
In this work, we present \acronym, a dataset of over 1 million podcast transcripts alongside foundational analyses for the continued study of the podcast ecosystem. 
Our data is comprehensive, characterizing this medium in terms of its content, audio characteristics, speaker roles, and a number of other relevant factors such as category, hosting service, and publication date. 
Our structural analysis describes what content is discussed by which categories and demonstrates that podcast communities are formed through shared guests. Our temporal analysis finds that the ecosystem responded similarly to how the news media respond to major events. Furthermore, discussion of George Floyd was widespread, reaching a diverse set of categories and a large fraction of total podcasts. 
Our data and analyses answer important questions about podcasts. More importantly, however, they open the door to the continued research of this impactful and understudied medium.

\section*{Limitations}

Our data and analysis are limited in a number of ways. With regard to our dataset, there are certain podcasts that are not publicly available and therefore excluded from our data collection pipeline. One noteworthy example is the popular Joe Rogan podcast, which was, until recently, available exclusively through Spotify.  Moreover, despite being a large online repository, Podcast Index is not guaranteed to have comprehensive coverage of existing podcasts. 

Our data processing pipeline involves additional limitations. Whisper models are predictive tools, meaning that they are inherently subject to mistakes and even hallucinations. These issues may be exacerbated in the context of non-majority accents, code-switching, or audio recorded with low-quality equipment or in noisy environments. In order to maximize data quality, we remove cases of potential hallucination, but this may disproportionately affect already marginalized groups. Furthermore, our diarization and role annotation approaches are subject to error that is difficult to quantify and could propagate to downstream analyses in unforeseen ways. In addition, some hosts do not introduce themselves, hence we are unable to identify them, and we do not attempt to resolve names to speakers in the case of more than one host or guest. 

While we provide a first large-scale analysis of podcasts, our conclusions are limited. In our analyses, we use a subset of data that is dense but represents a narrow slice of time. Thus, our conclusions may not generalize outside this timeframe.
Further limitations with regard to our temporal analysis stem from our inability to pinpoint mechanistic factors underlying our observed time series. While prior work is suggestive of particular explanations for the patterns observed within and across categories, our data do not permit a definitive answer for ``why'' these patterns emerge.

\section*{Ethics Statement}
The processing and open release of our data comes with potential harms. We have carefully considered the ethical implications of these harms, seeking to minimize their impact while maximizing potential benefits.  

At the data processing phase, one potential harm arises if a speaker's actual speech is mistranscribed or misattributed. Such a mistake would be particularly problematic if this speech is offensive or reflects stereotypes or other biases. Furthermore, this type of error could be more likely for users with non-majority ways of speaking, which could exacerbate this harm for groups that are already marginalized. To mitigate this issue at the transcription level, we have undertaken thorough manual review to ensure that hallucinations are rare and that repeated text is filtered from our data. In the case of diarization, this issue is more difficult to mitigate. Recognizing that the potential for harm due to speaker misattribution varies widely across contexts, we encourage future users of our data to consider how this potential error could manifest when interpreting downstream results. 

We further identify two potential harms due to the open release of our data. First, releasing our data could result in privacy concerns. Although we only process podcasts that are publicly available, the release of our processed data could expose a podcast's content to an unforeseen or undesired audience. Creators may have expectations that their content will not be widely distributed, and the release of our data could violate this expectation. Furthermore, the enhanced searchability afforded by podcast transcription could render a podcast's content easier to access than was originally intended. A second and related harm is the usage of data for training of large language models. Some creators may not want their content being used to train deep learning systems, and the release of our data makes content more accessible for these purposes. This issue is particularly concerning when content is used to generate revenue without compensating the original creators. 

To mitigate the ethical concerns of releasing our data, we limit  access in two ways. First, we allow podcast creators to request that their data be removed such that future downloads of the dataset no longer contain their content. Next, we limit the use of our dataset to
research purposes only, and will require users to acknowledge these limitations and intended use cases in order to gain access to the data. %

\section*{Acknowledgments}

This work was supported in part by the
National Science Foundation under Grant No. IIS-2143529 and by DSO National Laboratories.

\bibliography{anthology,custom}

\clearpage 

\appendix

\section*{Appendix}

\section{Transcription}
\label{app:transcription}

\paragraph{Details}
After preliminary experiments with different model sizes, we settled on using the \texttt{base.en} version of Whisper, with 74M parameters, as a compromise between speed and accuracy. To enable highly parallelized throughput, we use a version adapted to run on CPUs.\footnote{\url{https://github.com/ggerganov/whisper.cpp}} Altogether, we transcribed over 650K hours of audio, which required approximately roughly 220K CPU hours.

Following manual review, we discover that Whisper sometimes generates phrases that are repeated many times in a row, especially when it encounters periods of silence, music, or code-switching between languages. To filter these out, we use 4-grams to quantify repeated text. If a podcast episode contains a single 4-gram whose frequency represents over 5\% of the total 4-gram frequency in the transcript, this episode is removed. This process removes roughly 11\% of our data, leaving 1.1 million podcast transcripts for processing.

\paragraph{Validation}
To help quantify the performance of Whisper on podcasts, we compare the output on a small sample of episodes for which we can obtain professional transcripts shared by podcasts creators. Through web searches, we identify six shows that release transcripts for every episode on their websites, complete with a named speaker for each speaking turn. Of these shows, one (\emph{Welcome to Night Vale}) typically consists of monologues, read by one speaker; another (\emph{This American Life}) often involves many speakers over the course of one episode; and the remaining four are typically extended one-on-one interviews.
From these shows, we download the transcripts for six episodes per show, starting from the beginning of May 2020, and taking each episode after that in turn. 

Unfortunately, we cannot directly compare the output of Whisper to these transcripts, in part because the transcripts often exclude things like the intro and outro to the episode, and may be annotated with additional text throughout (including speaker names). Thus, we first process the transcript files to remove everything but the spoken words. We then manually edit the output of Whisper to remove everything before the first word and after the last word of the transcript.\footnote{In none of these cases did we encounter hallucinated passages that were not in the audio files.}

\begin{figure}[t]
    \centering
    \includegraphics[width=1\linewidth]{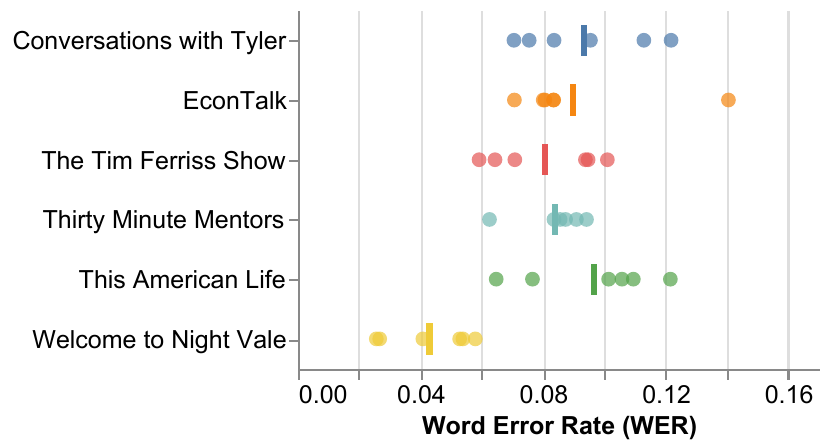}
    \caption{Estimated word error rates from Whisper, based on comparison to professional transcripts for six episodes from each of six shows. Vertical bars show averages across episodes. The low error rates for the performed monologues featured on \emph{Welcome to Night Vale} hint at the fact that many of these apparent errors are actually due to the way most professional transcripts are edited to remove disfluencies in speech (see below).
    }
    \label{fig:auto_wer}
\end{figure}

Given these preprocessed versions, for each episode, we tokenize both the transcript and the whisper output using spaCy, drop punctuation, and then use NLTK to align the two resulting lists of tokens based on edit distance. To compute word error rate (WER), we add up the number of insertions, deletions, and errors, and divide by the total number of tokens in the aligned list. 

Our estimated WER values are show in Figure \ref{fig:auto_wer}, with average WER per show shown by vertical bars. In most cases, the average WER is between 8-9\%, which is better than the average WER reported in the original Whisper paper.\footnote{The Whisper paper \citep{radford2022robustspeechrecognitionlargescale} reported an average WER of 12.8\% for Whisper Large V2 across 13 datasets, with average WER per dataset varying between 3.9\% and 36.4\%. } However, the error rates measured by this fully automated comparison actually overstate the true error rate by a considerable degree. The reason is that the professionally produced transcripts are both imperfect, and intentionally clean up the words spoken on the episodes, often removing disfluencies, like filler words (e.g., ``um'', ``like'', ``you know'', etc.) and repetitions (e.g., ``I mean, I mean''). Whisper does sometimes elide these as well, but is generally much closer to an exact transcription. The effect of this can be seen in the low WER for \emph{Welcome to Night Vale} (average WER of 4.3\%), which consists of professional readings of pre-written scripts \citep{mcglynn.2014.interview}, and thus contains far fewer of these disfluencies than natural speech.%

To get a better estimate of the true transcription performance of Whisper, we manually annotate the alignment files for the first five minutes (from the start of the transcript) for each of eighteen episodes (the first three episodes from each show in our validation sample), and mark cases where Whisper was legitimately correct or incorrect, and with what kind of error. In doing these annotations, we are strict with respect to words being correct (so that ``Amazons'' is \emph{not} equivalent to ``Amazon's''), but we do count equally plausible spellings of words and phrases as correct (except for proper names). This includes numbers (e.g., ``11'' vs. ``eleven''), times (e.g,. ``9:00pm'' vs. ``9 o'clock p.m''), money (e.g., ``\$279'' vs. ``two hundred seventy nine bucks''), alternative spellings (e.g., ``advisor'' vs. ``adviser''). This also includes words that could plausibly be split (e.g., ``best selling'' vs. ``bestselling''), and conjunctions and alternatives which cannot be easily decided from the audio (e.g,. ``Today is'' vs. ``Today's'' or ``going to'' vs. ``gonna'').

\begin{figure}[t]
    \centering
    \includegraphics[width=1\linewidth]{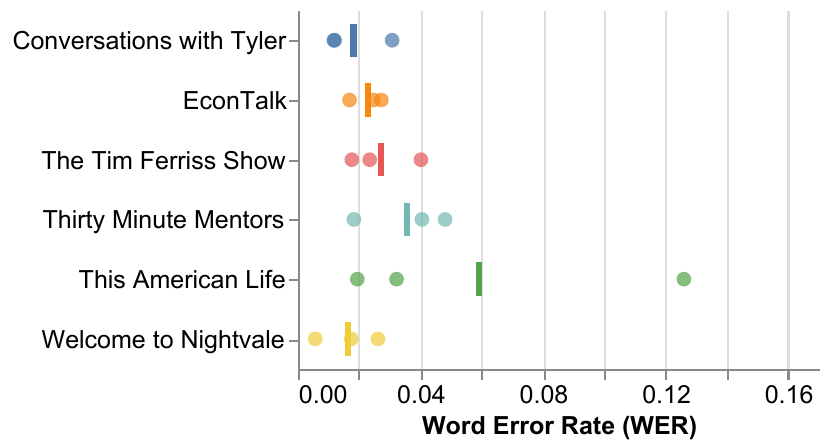}
    \caption{Manually estimated word error rates from Whisper, based on listening to the first five content minutes of three episodes from each of six shows. Vertical bars show averages across episodes. The high error rate for one episode of \emph{This American Life} is primarily due to Whisper's failure to transcribe the Spanish words in that episode.
    }
    \label{fig:whisper_manual_wer}
\end{figure}

The results of this manual analysis are presented in Figure \ref{fig:whisper_manual_wer}, and reveal that Whisper is actually performing much better than our automated analysis initially suggested. Based on our manual analysis of these eighteen episodes, we compute an average WER of 3.0\%. However, it turns out that this even number is inflated somewhat by one particular episode of \emph{This American Life}, which contains a high number of Spanish words that Whisper fails to translate, and which thus ends up with a WER of 12.6\%. If we exclude this episode, then our manual analysis produces an estimated average WER for English of 2.4\%, which is just under what the Whisper paper reported as performance on LibriSpeech Clean (2.7\%).\footnote{Note, however, that just as our automated analysis was too pessimistic, our manual analysis may be slightly too optimistic. The reason is that we do not penalize Whisper for omitting filler words or disfluencies that the professional transcripts also omit. These were quite rare, but including these would slightly increase the WER, depending on what we consider to be the ``correct'' transcription. Ultimately, mapping conversational speech to a clean sequence of tokens is a complex task, and may not be well-defined if only working at the level of tokens, especially in the case of overlapping speech.}

Looking at the annotated alignments in more detail, we find that in cases where there is a disagreement about a word between Whisper and the professional transcript, we find that the Whisper is correct about half the time. In cases where Whisper produces the wrong word, this is most commonly due to proper names, which are legitimately ambiguous in terms of spelling, without additional world knowledge. In cases where the transcript is missing a word that is in the Whisper output, Whisper is correct that the word was spoken more than 80\% of the time. When Whisper is missing a word that is in the transcript (which is the least common type of error), this is usually a legitimate error, where Whisper has failed to transcribe a word, which is often a filler word that was retained by the transcript.

Without further analysis, it is difficult to know how well these estimates will generalize to other shows. However, the fact that we observe error rates of around 4\% or less on most episodes, across six shows with some diversity of speakers, suggests that transcription performance tends to be quite reliable overall.

\section{Diarization}
\label{app:diarization}

\paragraph{Details}
We perform diarization using \texttt{pyannote} on a mix of GPUs (A5000 and A6000s). Overall, we diarize more than 220K hours of audio, which required approximately 22K GPU hours. Through a through manual review of speaker labels assigned by \texttt{pyannote}, we found that it identified changes between speakers with high accuracy; however, some short segments of audio may be assigned to extraneous speaker labels, resulting in extra speaker labels with only a small amount of short segments assigned to them. To address this issue, we filter out speakers whose total speaking time is less than 5\% of the total speaking time in an episode. This filtering process also helps to remove advertisements and short introductions by speakers who are not actually present within the main podcast content. 

\paragraph{Validation}
To validate the performance of \texttt{pyannote} on this data, we again make use of the professional transcripts described in Appendix \ref{app:transcription}. Working with the raw diarization output (prior to filtering), we similarly manually edit these to remove everything before the first word of the transcript and everything after the last word of the transcript. For the preprocessed version of each of the professional transcript and the pyannote output, we again tokenize with spaCy, drop punctuation, and convert each to a sequence of tuples, where each tuple represents one token and the corresponding set of one or more speakers. For the professional transcripts, this is always a single named speaker per token. For the diarization output the speakers are unnamed (e.g., speaker$_1$, speaker$_2$), and the set might contain one speaker or more (in the case of inferred overlap). 

To determine how well the two assignments of tokens to speakers align, we rely on the token alignments created during our transcription validation (see Appendix \ref{app:transcription}). In particular, we use the alignment between the whisper output and the professional transcript to get the named speaker from the transcript for each token.\footnote{Because of minor differences in tokenization which sometimes happen, we allow for matching sequences of tokens (rather than just individual tokens) as necessary.} For each token in the diarization output, we thus have the (correct) named speaker, and a set of one or more unnamed speakers. In doing so, we produce co-occurrence statistics on how many times each named speaker from the professional transcript is matched to each unnamed speaker from diarization. Finally, we associated each named speaker with the one unnamed speaker with the highest number of associated tokens, who has not yet been assigned to a name speaker. We do this in order of most tokens spoken among the named speakers.\footnote{Note that this means that in some cases, an unnamed speaker from diarization might not get matched to any named speaker from the transcript.}

Using this alignment of one unnamed speaker from diarization for each named speaker from professional transcripts, we can thus determine the proportion of tokens that have been correctly assigned. We compute the diarization token error rate for each episode, where we count as an error any token assigned to multiple unnamed speakers, or assigned to the wrong unnamed speaker (i.e., anyone other than the one associated with the correct named speaker for that token). 

\begin{figure}[t]
    \centering
    \includegraphics[width=1\linewidth]{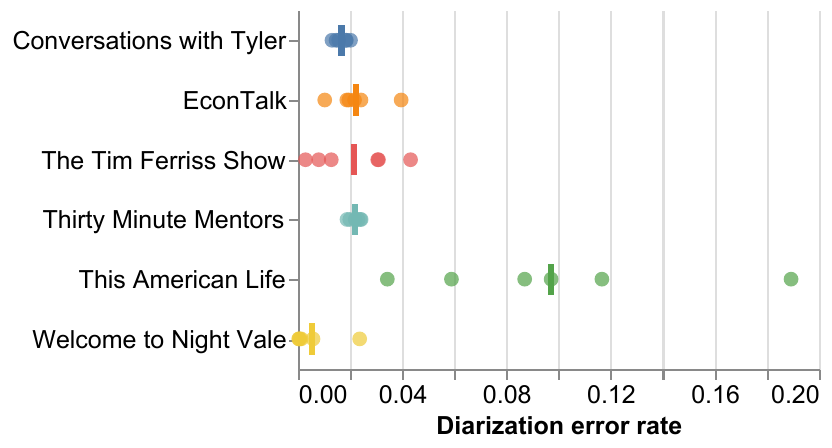}
    \caption{Estimated token error rates for diarization, again based on comparison to professional transcripts for six episodes from each of six shows. Vertical bars show average error rates across episodes. The higher error rates for  \emph{This American Life} results from the more challenge setting of having many more speakers per episode.
    }
    \label{fig:diarization_error_rate}
\end{figure}

The results of this are highly encouraging, as shown in Figure \ref{fig:diarization_error_rate}. With the exception of one show, the diarization error rate for all episodes is less than 5\%, and for most episodes less than 2\%. The one show for which diarization struggles more (\emph{This American Life}), is much more challenging due to the higher number of speakers on each episode. In our sample of six episodes, all episodes featured at least 17 speakers, with an average of 23.5 per episode. Note that although \emph{Welcome to Night Vale} typically features only a single speaker, getting less than 100\% diarization accuracy is possible if \texttt{pyannote} incorrectly infers that there is more than one speaker present in the audio. If we exclude \emph{This American Life}, the overall average diarization token error rate is 1.8\%, or 2.1\% for those episodes which feature one-on-one interviews.

\section{Speaker Role Labeling}
\label{app:role-labeling}

Because hosts and guests are not identified in any systematic fashion, we developed a system to determine this automatically, where possible. In particular, we rely on the fact that hosts frequently identify themselves and their guests at the start of an episode. We begin by identifying candidate names by using spaCy to perform NER on the first 350 words of the transcript and in podcast and episode descriptions. We keep only \textsc{Person} entities, and only those that are two-word phrase (rather than single names). 
We further note that episodes can have one or most hosts, and zero or more guests.
Based on this setup, we develop and deploy a system to classify each extracted name as \textsc{Host}, \textsc{Guest}, or \textsc{Neither}.

\paragraph{Annotation}
In order to train a model to perform this classification, we collect human judgments as to whether extracted names are \textsc{Host}, \textsc{Guest}, or \textsc{Neither}. We set this up as a classification task for humans, in which we show annotators a single target entity name, along with the necessary context (podcast description, episode description, and 150 words surrounding the name's first appearance in the transcript).
Using Prolific and the Potato annotation tool \citep{pei2022potato}, we collect three human judgments on each of 2,000 entity names.
The task consists of a sequence of categorical annotation questions, each of which shows occurrences of the same named entity within podcast descriptions, episode descriptions, and episode transcripts. 

\begin{figure*}[hbt!]
\centering
\includegraphics[width=\textwidth]{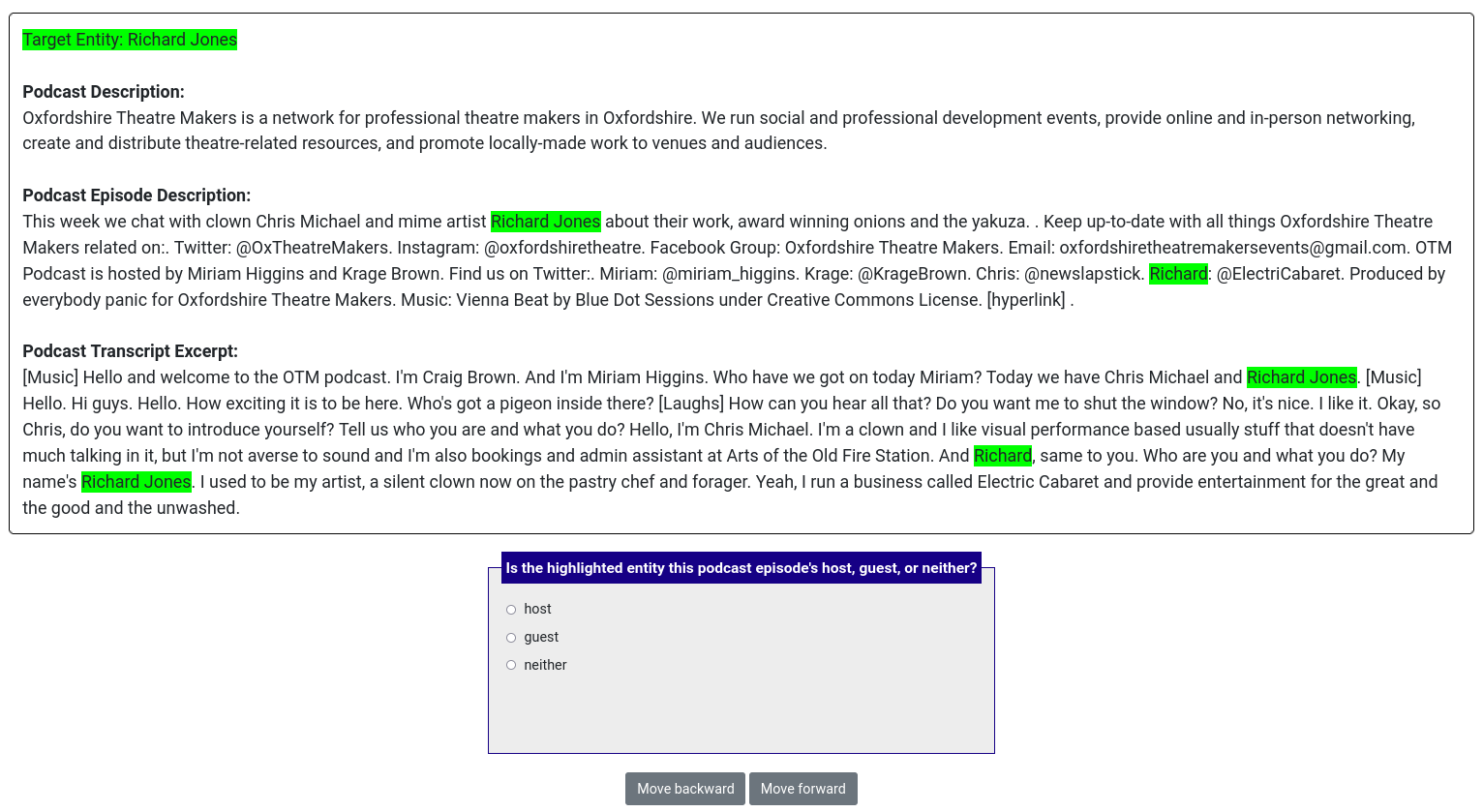}
\caption{A screenshot of our annotation task using the Potato tool. Note that a full 300 word window of transcript text does not appear, since the first named entity occurs close to the beginning of the transcript. }
\label{fig:annotationSS}
\end{figure*}

Our annotators were drawn from a pool of roughly 40K individuals on Prolific with a Bachelor's or Associate's degree who speak English as their first language. Participants were given batches of 43 entities to be labeled, with 3 of these entities being sanity checks with a simple and correct answer. Estimated time for this task was 16 minutes. The pay for task completion was 4 dollars and participants were paid roughly 13 dollar per hour on average. 

We carry out our annotation over two rounds.
To select our 1,000 examples for the first round of annotation, we chose a stratified random sample of 50 unique entity names from each of the top 20 podcast categories. This process ensured a balance of different types of entities occurring in a variety of contexts, as well as ensuring that very common entity names (e.g. George Floyd) did not dominate our annotation set. 

For our second round of annotation, we trained an initial host-guest model on our first round annotations and took the 1,000 entities with the lowest confidence predictions from a random sample of 15,000 unique entity predictions.

Our annotation pipeline included 3 annotations per entity. Chance corrected annotator agreement (using Krippendorff's $\alpha$) was  0.77 for the first round, and 0.53 for the second round, since the second round consisted of more difficult examples. To account for malicious annotation strategies (e.g. spamming), we used the MACE tool to predict the most probable annotation for each entity \cite{hovy2013learning}. 

\begin{figure*}[hbt!]
\centering
\includegraphics[width=\textwidth]{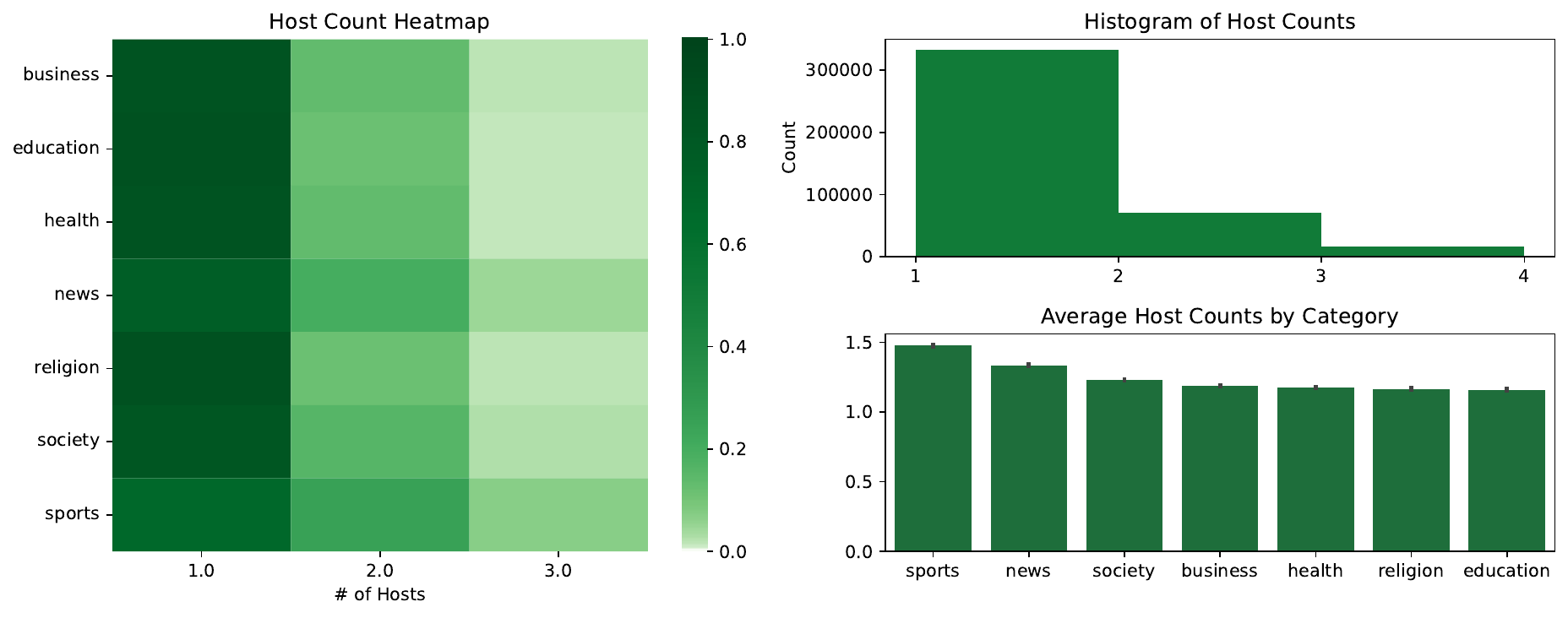}
\caption{Our role annotation model most commonly identifies the names of one host, however, the rate of host identification differs across categories. Our Host Count Heatmap (left) represents the distribution of predicted guests across different categories. These distributions are further summarized by the Average \textsc{Host} counts by Category (lower right) which shows that \textsc{sports}, \textsc{news}, and \textsc{society} have the most inferred \textsc{Hosts} on average. Finally we include our Histogram of \textsc{Host} Counts (upper right) which shows how many times a particular number of hosts was inferred (i.e. 1 host was inferred $\sim$300K times). A square root transformation was applied to the data in our heatmap to help better visually distinguish differences in the tails of these distributions across categories. Although no hosts are inferred for many podcasts, we exclude them from our summary here, assuming that all podcasts have a host. These cases are likely those where hosts did not identify themselves at all or did so only with their first name.}
\label{fig:hostCount}
\end{figure*}

\begin{figure*}[hbt!]
\centering
\includegraphics[width=\textwidth]{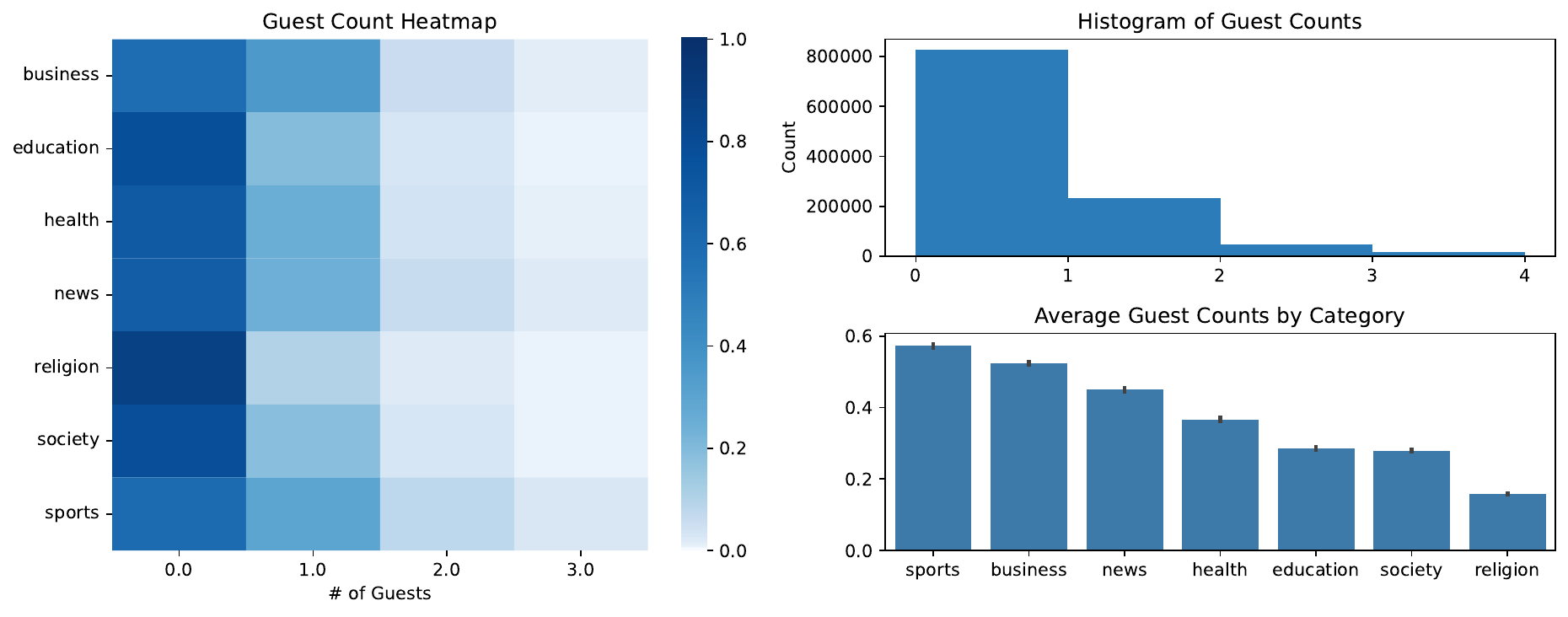}
\caption{Our role annotation model most commonly identifies the names of zero \textsc{Guests}, however, the rate of guest identification differs drastically across categories. Our Guest Count Heatmap (left) represents the distribution of predicted guests across different categories. These distributions are further summarized by the Average Guest Counts by Category (lower right) which shows that \textsc{sports}, \textsc{business}, and \textsc{news} have the most inferred Guests on Average. Finally we include our Histogram of Guest Counts (upper right) which shows how many times a particular number of guests was inferred (i.e. 1 guest was inferred $\sim$200K times). A square root transformation was applied to the data in our heatmap to help better visually distinguish differences in the tails of these distributions across categories.}
\label{fig:guestCount}
\end{figure*}

\paragraph{Model Training}
Our role annotation model consists of a RoBERTa model with an additional linear layer of dimension (512 x 3) and subsequent softmax transformation. For each entity occurrence in a transcript, the entity is fed into the model with 50 words on each side for context. The model was trained with a learning rate of $2e\textrm{-}6$, a batch size of 4, for 5 total epochs. Although we annotate 2,000 total entities, some entities occurred multiple times within the first 350 words of transcripts. In these cases we use the prediction with the max probability. 

\paragraph{Evaluation and inference} 
To identify the names of hosts and guests for each episode, we apply our trained model to all the extracted entity names for that episode.
Once again, the maximum probability prediction was used for entities occurring multiple times in the same transcript. To evaluate the model, we only used entities from the first round of annotation in our held out set, since they are a more representative sample. %
Our final model achieved a mean cross validation accuracy of .87 and a test accuracy of .88. Distributions of the number of inferred hosts and guests across categories are shown in Figures \ref{fig:hostCount} and \ref{fig:guestCount}.

\section{Data Merging and Dataset Release}
\label{app:dataset}

As a first step in creating our final data for release, we merge transcription, prosodic, and diarization data into a single file format. This is done at the word-level which involves assigning each tenth-of-a-second window of prosodic information to the word with maximal temporal overlap and aggregating.
Finally, speaker labels from diarization are assigned to each word, with cases of overlapping speaker assignment maintained in the data.

We release our dataset in two formats corresponding to podcast and speaker-turn level information respectively. At the podcast level, we release full transcripts, podcast and episode level metadata, and inferred host and guest assignments. At the speaker-turn level, we provide aggregated prosodic features, inferred speaker roles and names, the transcribed speaker turn, and a unique key mapping the turn back to the podcast-level metadata. When identifying speaker turns for segments of words with overlapping speaker labels, we assign this text to the speaker who had already been speaking when the overlap occurred. %

\section{Descriptive Overview}
\label{descriptiveOverview}

\begin{figure*}[hbt!]
\centering
\includegraphics[width=\textwidth]{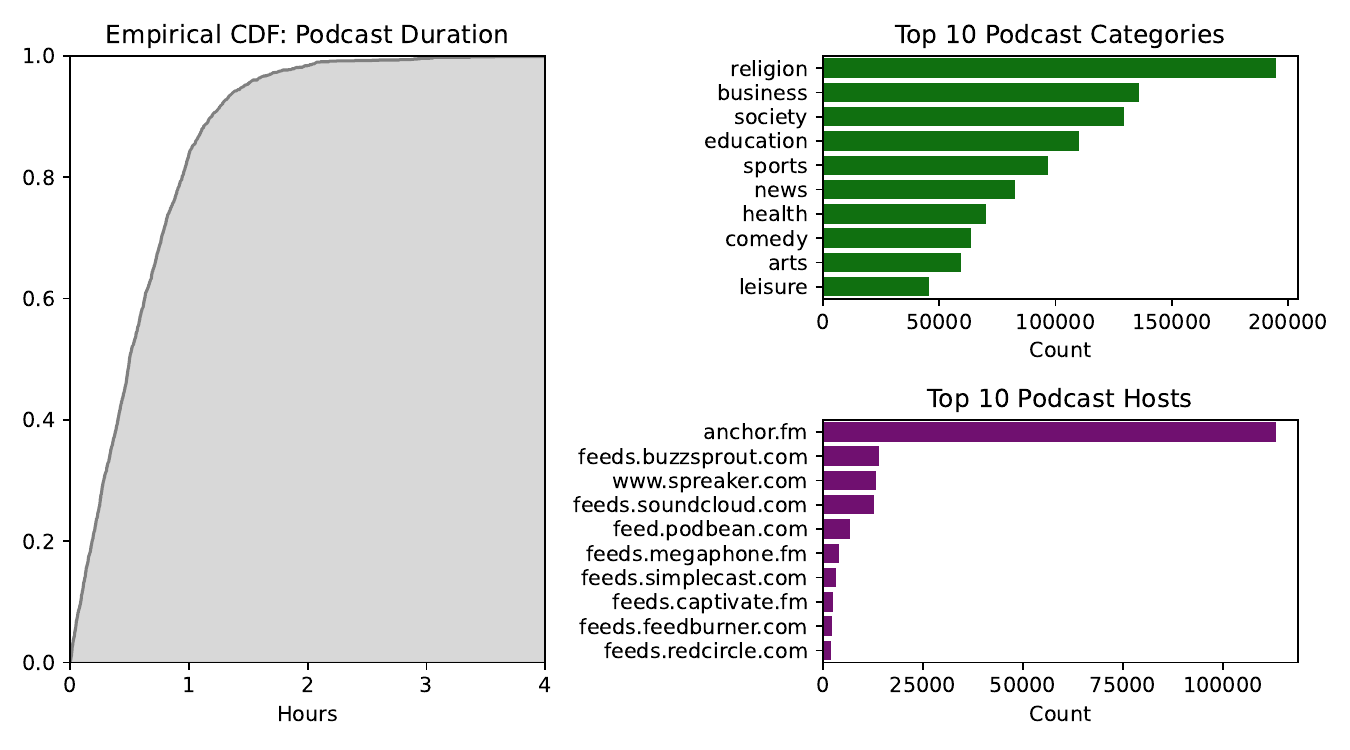}
\caption{Podcasts fall into a number of top categories, are dominated by a single platform, and have a median duration of 30 minutes. The upper right plot shows the total number of episodes in our dataset for the top 10 podcast categories. The lower right figure shows the counts of podcast hosting platforms in our metadata, where Spotify's hosting platform anchor is by far the largest. On the left, we plot an empirical CDF of podcast durations that have been extracted from our metadata and formatted.}
\label{fig:durationCatHost}
\end{figure*}

Figure \ref{fig:durationCatHost} summarizes statistics of our podcast dataset regarding length, hosting, and categories. We find a median podcast length of 30 minutes, with approximately 80 percent of podcasts lasting one hour or less. Our data also show that podcast hosting is highly centralized, with \texttt{anchor.fm} (owned by Spotify) hosting nearly 6 times the number of podcasts as SoundCloud, the next largest host. This centralization is in line with prior work regarding the medium's evolution as a source for corporate profit and mass media communications \cite{bonini.2015.second}. Finally, podcast creators can assign their podcasts to a number of categories. We find that \textsc{Religion} is by far the largest category, followed by \textsc{Society}, \textsc{Education}, \textsc{Business}, \textsc{Sports}, and \textsc{News}.

\section{Speech Characteristics}

Although it is not the focus of our work, we provide one example of the potential uses of extracted audio information. Using the fundamental frequency (F0) values from OpenSMILE, we compute the average pitch per category of podcasts in our sample.

\begin{figure}
    \centering
    \includegraphics[width=0.95\linewidth]{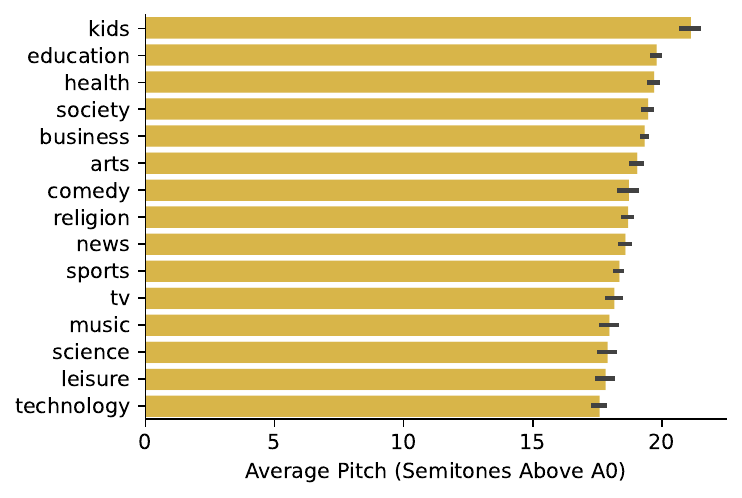}
    \caption{Average fundamental frequency (F0) values (in semitones above A0) by category, with error bars showing 95\% confidence intervals.}
    \label{fig:pitch-by-category}
\end{figure}

As expected, podcasts in the \textsc{Kids} category (aimed at children) have the highest average pitch. By contrast, we find the lowest average pitch in \textsc{Science}, \textsc{Leisure}, and \textsc{Technology} categories, a pattern which may reflect a relative lack of gender diversity (see  Figure \ref{fig:pitch-by-category}). Future analyses could look at variation within these categories, and other speech properties such as dialect and speaking rate, as well as additional vocal characteristics measured by OpenSMILE.

\section{Host-Guest Network}
\label{app:network}

Constructing the network of guest co-appearances across podcasts requires identifying unique guests that appear on multiple shows. Unfortunately, based only on names, there is the potential for misidentifying all individuals with the same name as one person.
To minimize the chance of this possibility, we filter out guests with common names.
We first estimate the probability of first and last names in our data separately, by splitting all identified named entities into two names (first and last), and computing the frequency of occurrence of each of these individually.
We then model the overall probability of a two-word entity name as $p(\textrm{name}) = p(\textrm{first})*p(\textrm{last})$. This naively assumes that first and last name probabilities are independent, which they clearly are not, but is sufficient for our purposes. We then keep only those guests having $p(\textrm{name})$ below the median name probability across all guests. 

\section{Calculating Modularity}
\label{app:modularity}

For a set of nodes partitioned into communities in a graph, modularity measures the difference between the number of edges that fall within their community minus the number that would be expected if the edges were reassigned randomly. More formally, modularity is given by equation (\ref{eq:modularity}), where $A_{ij}$ indicates whether two nodes are connected based on adjacency matrix $A$, $\frac{k_i k_j}{2m}$ is the (approximate) expected probability of an edge occurring between nodes $i$ and $j$, and $\delta(c_i, c_j)$ is an indicator variable for whether $i$ and $j$ are in the same community \cite{newman2006}.  

\begin{equation}
    Q = \frac{1}{2m} \sum_{i,j} \left[ A_{ij} - \frac{k_i k_j}{2m} \right] \delta(c_i, c_j)
    \label{eq:modularity}
\end{equation}

For each category $c$ we create a new partition of the network into nodes in $c$ and nodes not in $c$. Modularity is then recalculated over these binary partitions to arrive at our results in table \ref{tab:modularity}. 

\section{Topics}
\label{app:topics}

The racial justice topics shown in Figure \ref{fig:2panelFloydSummary} in the main paper are provided in Table \ref{tab:racial-justice-topics} below.

\begin{table}[h!]
    \centering
    \small
    \begin{tabular}{l l}
        \textbf{Topic} & \textbf{Top words} \\
        \hline
        George Floyd & people, george, black, floyd, police  \\
        BLM & black, lives, matter, people, racism \\
        Policing & police, officers, officer, law, protests \\
        Race & black, white, people, color, racism \\
        Society & social, people, culture, society, change \\
        Am. History & american, history, states, united
    \end{tabular}
    \caption{Racial justice topics plotted in Figure \ref{fig:2panelFloydSummary}.}
    \label{tab:racial-justice-topics}
\end{table}

A full detailed list of topics are presented in Table \ref{tab:allstorms}, sorted by overall prevalence. Entropy is calculated in terms of the proportions across categories. Relative Entropy is calculated with respect to the overall distribution of categories, and thus captures how concentrated topics are within categories. 

\section{Racial Justice Topics Across Categories}
\label{racialJusticeAppendix}

\begin{figure*}[h!]
\centering
\includegraphics[width=\textwidth]{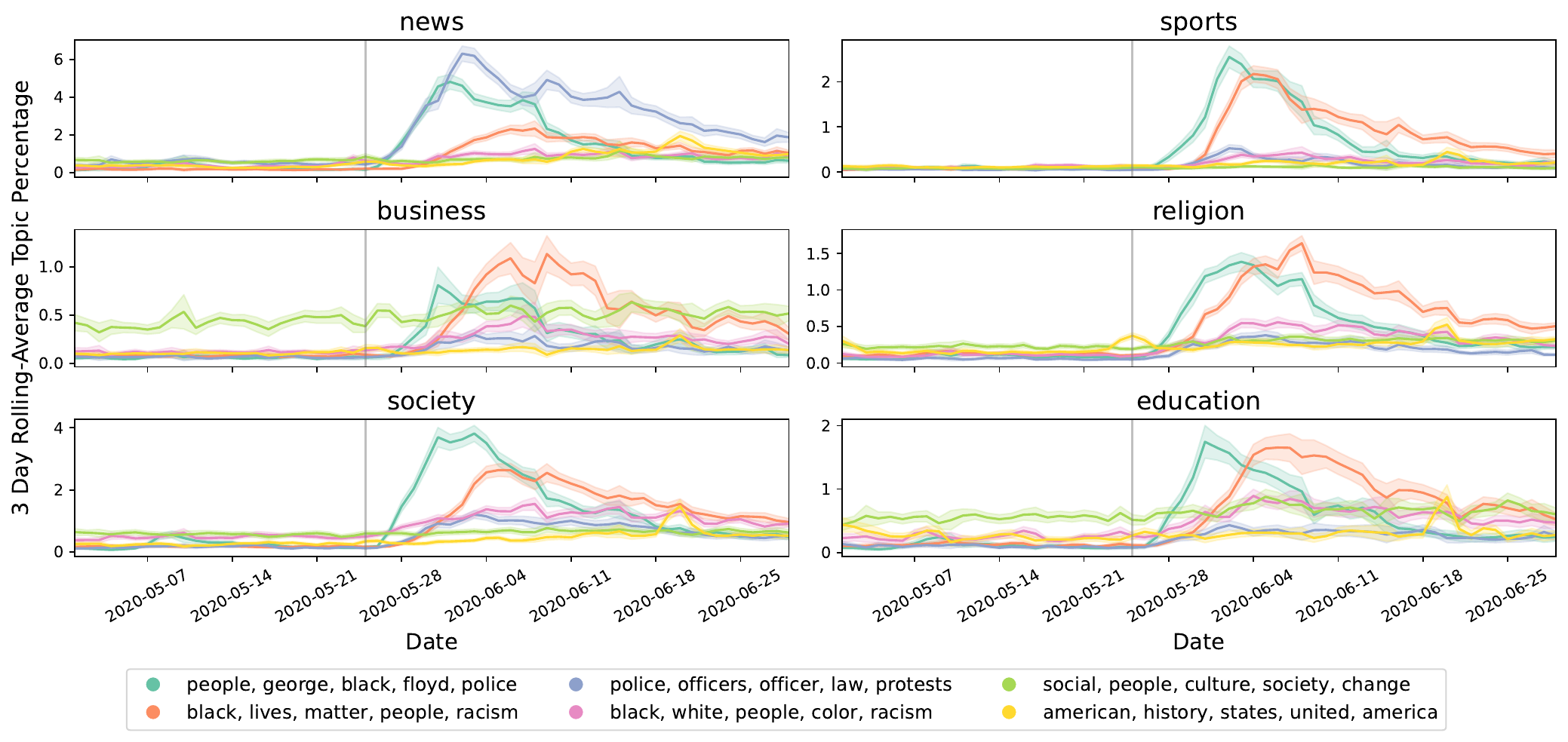}
\caption{The temporal lag between ``Black, Lives, Matter'' and ``People, George, Floyd'' is consistent across most categories, however News devotes far more coverage proportionally to ``Police, Officers, Law'' and Business appears to devote a larger share of its content to Black Lives Matter than George Floyd.}
\label{fig:categoryTimeSeries}
\end{figure*}

Variations on Figure \ref{fig:2panelFloydSummary}, with the rise and fall of racial justice topics broken down by category for each of the top categories is shown in Figure \ref{fig:categoryTimeSeries}. As can be seen \textsc{News} differs dramatically from other categories in focusing much more on policing and protests, whereas others focus much more on George Floyd and Black Lives Matter.

\centering
\begin{tiny}
\onecolumn
\begin{longtable}{L{3.7 cm}  R{1 cm}  R{.8 cm}  R{.8 cm} L{4cm} L{3cm}}
\label{topicTable}
Top Words & Topic Proportion & Entropy & Relative Entropy & Top Categories & Top Category Proportions \\
\hline
we're it's that's things talk & 0.157 & 2.433 & 0.041 & 'religion', 'business', 'society', 'education' & 0.183, 0.169, 0.125, 0.112 \\
didn't time back started thought & 0.152 & 2.388 & 0.119 & 'society', 'religion', 'education', 'health' & 0.195, 0.165, 0.141, 0.115 \\
you're don't it's i'm that's & 0.147 & 2.416 & 0.111 & 'education', 'society', 'religion', 'business' & 0.172, 0.168, 0.126, 0.123 \\
it's i'm i've that's good & 0.125 & 2.553 & 0.095 & 'society', 'education', 'business', 'comedy' & 0.184, 0.1, 0.094, 0.088 \\
people don't it's lot things & 0.119 & 2.271 & 0.176 & 'society', 'education', 'religion', 'business' & 0.254, 0.186, 0.115, 0.102 \\
kind it's lot things bit & 0.117 & 2.570 & 0.095 & 'society', 'business', 'education', 'health' & 0.162, 0.127, 0.102, 0.092 \\
episode podcast i'm talk episodes & 0.108 & 2.561 & 0.126 & 'society', 'education', 'business', 'comedy' & 0.203, 0.133, 0.085, 0.075 \\
it's they're people that's there's & 0.107 & 2.520 & 0.072 & 'society', 'news', 'religion', 'comedy' & 0.169, 0.126, 0.111, 0.101 \\
yeah it's that's good don't & 0.103 & 2.497 & 0.313 & 'comedy', 'society', 'tv', 'leisure' & 0.201, 0.173, 0.089, 0.089 \\
yeah it's i'm don't that's & 0.090 & 2.436 & 0.410 & 'comedy', 'society', 'leisure', 'tv' & 0.218, 0.166, 0.117, 0.103 \\
years started job work working & 0.078 & 2.128 & 0.359 & 'business', 'education', 'society', 'health' & 0.371, 0.139, 0.125, 0.071 \\
day week good morning today & 0.078 & 2.522 & 0.090 & 'society', 'news', 'education', 'sports' & 0.168, 0.121, 0.116, 0.095 \\
sort it's that's there's bit & 0.076 & 2.584 & 0.109 & 'business', 'society', 'arts', 'education' & 0.183, 0.116, 0.089, 0.085 \\
home work time people it's & 0.075 & 2.467 & 0.107 & 'society', 'business', 'education', 'health' & 0.185, 0.172, 0.124, 0.08 \\
she's i'm yeah love it's & 0.073 & 2.528 & 0.190 & 'society', 'comedy', 'tv', 'arts' & 0.192, 0.095, 0.093, 0.091 \\
gonna i'm wanna we're upbeat & 0.071 & 2.485 & 0.113 & 'society', 'education', 'comedy', 'business' & 0.196, 0.138, 0.102, 0.099 \\
guys i'm we're it's yeah & 0.068 & 2.574 & 0.111 & 'sports', 'society', 'business', 'leisure' & 0.145, 0.126, 0.113, 0.087 \\
time fact sense man human & 0.067 & 2.188 & 0.335 & 'religion', 'arts', 'education', 'society' & 0.29, 0.173, 0.141, 0.128 \\
number it's that's let's there's & 0.065 & 2.387 & 0.257 & 'education', 'business', 'society', 'religion' & 0.277, 0.147, 0.12, 0.061 \\
he's yeah guy i'm that's & 0.065 & 2.293 & 0.564 & 'comedy', 'sports', 'society', 'tv' & 0.315, 0.125, 0.102, 0.093 \\
i'm don't it's that's you're & 0.065 & 2.351 & 0.532 & 'comedy', 'society', 'tv', 'leisure' & 0.314, 0.16, 0.066, 0.066 \\
year week back month june & 0.065 & 2.555 & 0.094 & 'society', 'news', 'sports', 'business' & 0.136, 0.12, 0.116, 0.106 \\
he's john guy tom great & 0.064 & 2.545 & 0.332 & 'tv', 'society', 'comedy', 'arts' & 0.146, 0.129, 0.116, 0.098 \\
we're show good i'm chris & 0.064 & 2.567 & 0.062 & 'business', 'sports', 'religion', 'society' & 0.129, 0.126, 0.11, 0.108 \\
yeah i'm we're show he's & 0.062 & 2.559 & 0.120 & 'sports', 'business', 'comedy', 'society' & 0.165, 0.146, 0.097, 0.096 \\
story stories podcast share people & 0.061 & 2.338 & 0.180 & 'society', 'business', 'education', 'religion' & 0.216, 0.159, 0.156, 0.111 \\
show free podcast website email & 0.058 & 2.675 & 0.079 & 'business', 'education', 'society', 'health' & 0.146, 0.12, 0.108, 0.086 \\
family dad mom years life & 0.056 & 2.364 & 0.196 & 'society', 'religion', 'education', 'business' & 0.262, 0.148, 0.123, 0.098 \\
it's bit i've yeah i'm & 0.055 & 2.567 & 0.167 & 'sports', 'society', 'comedy', 'business' & 0.138, 0.134, 0.119, 0.104 \\
important process make information problem & 0.053 & 1.958 & 0.540 & 'business', 'education', 'health', 'society' & 0.324, 0.309, 0.06, 0.056 \\
money pay people make million & 0.052 & 2.232 & 0.220 & 'business', 'education', 'society', 'news' & 0.297, 0.159, 0.116, 0.097 \\
house room door home back & 0.051 & 2.586 & 0.211 & 'society', 'comedy', 'education', 'religion' & 0.168, 0.133, 0.099, 0.098 \\
world pandemic crisis time people & 0.049 & 2.227 & 0.209 & 'business', 'religion', 'education', 'news' & 0.2, 0.197, 0.136, 0.126 \\
it's feel life i'm experience & 0.049 & 1.946 & 0.449 & 'education', 'health', 'society', 'religion' & 0.272, 0.186, 0.169, 0.155 \\
york city california san texas & 0.049 & 2.428 & 0.172 & 'society', 'business', 'sports', 'arts' & 0.257, 0.128, 0.094, 0.086 \\
man i'm that's what's it's & 0.047 & 2.312 & 0.322 & 'society', 'comedy', 'sports', 'news' & 0.258, 0.121, 0.105, 0.098 \\
question questions answer asked number & 0.047 & 2.462 & 0.298 & 'leisure', 'education', 'religion', 'society' & 0.213, 0.122, 0.099, 0.097 \\
life success goal goals things & 0.046 & 1.777 & 0.570 & 'education', 'business', 'health', 'society' & 0.372, 0.251, 0.119, 0.09 \\
good day hope positive today & 0.045 & 2.032 & 0.383 & 'education', 'society', 'religion', 'health' & 0.32, 0.178, 0.132, 0.13 \\
time things day work you're & 0.044 & 2.013 & 0.450 & 'education', 'business', 'health', 'society' & 0.293, 0.222, 0.151, 0.098 \\
i'm i've bit i'll talk & 0.044 & 2.550 & 0.211 & 'society', 'comedy', 'sports', 'music' & 0.183, 0.148, 0.077, 0.072 \\
book books read reading author & 0.044 & 1.993 & 0.820 & 'arts', 'education', 'society', 'business' & 0.466, 0.096, 0.089, 0.084 \\
life live living world purpose & 0.044 & 1.855 & 0.431 & 'education', 'religion', 'society', 'health' & 0.311, 0.281, 0.126, 0.094 \\
social media facebook instagram people & 0.043 & 2.149 & 0.347 & 'business', 'society', 'education', 'news' & 0.374, 0.134, 0.119, 0.064 \\
change mind make things control & 0.043 & 1.856 & 0.513 & 'education', 'health', 'business', 'religion' & 0.353, 0.161, 0.153, 0.139 \\
i'm women life love today & 0.042 & 2.037 & 0.361 & 'business', 'education', 'society', 'health' & 0.234, 0.219, 0.155, 0.142 \\
school high year grade college & 0.041 & 2.423 & 0.125 & 'society', 'education', 'religion', 'business' & 0.212, 0.182, 0.091, 0.08 \\
music playing upbeat gentle soft & 0.041 & 2.546 & 0.212 & 'society', 'religion', 'music', 'education' & 0.156, 0.135, 0.126, 0.102 \\
video youtube channel videos live & 0.040 & 2.502 & 0.223 & 'business', 'leisure', 'society', 'education' & 0.187, 0.136, 0.131, 0.094 \\
podcast podcasts spotify apple listen & 0.040 & 2.611 & 0.124 & 'society', 'comedy', 'education', 'business' & 0.163, 0.119, 0.097, 0.088 \\
fire death life dream die & 0.039 & 2.276 & 0.395 & 'religion', 'arts', 'society', 'music' & 0.227, 0.164, 0.145, 0.115 \\
it's don't news blah people & 0.038 & 2.435 & 0.314 & 'news', 'society', 'comedy', 'sports' & 0.203, 0.191, 0.151, 0.062 \\
people truth good person wrong & 0.038 & 1.777 & 0.416 & 'religion', 'education', 'society', 'health' & 0.443, 0.175, 0.15, 0.069 \\
light tree sun birds beautiful & 0.038 & 2.442 & 0.474 & 'arts', 'religion', 'education', 'society' & 0.192, 0.139, 0.126, 0.123 \\
north weather carolina south west & 0.036 & 2.468 & 0.332 & 'society', 'news', 'sports', 'science' & 0.197, 0.182, 0.116, 0.083 \\
kids children child parents young & 0.036 & 2.104 & 0.781 & 'kids', 'education', 'religion', 'society' & 0.285, 0.176, 0.137, 0.13 \\
feel feeling emotions pain emotional & 0.035 & 1.906 & 0.478 & 'education', 'religion', 'health', 'society' & 0.235, 0.224, 0.214, 0.166 \\
love heart loved loving loves & 0.034 & 1.741 & 0.495 & 'religion', 'society', 'education', 'health' & 0.439, 0.169, 0.164, 0.052 \\
show radio news network live & 0.034 & 2.505 & 0.230 & 'news', 'society', 'business', 'religion' & 0.234, 0.12, 0.084, 0.077 \\
sleep morning night day bed & 0.034 & 2.413 & 0.311 & 'health', 'education', 'society', 'comedy' & 0.182, 0.16, 0.158, 0.09 \\
back side hand left feet & 0.034 & 2.394 & 0.377 & 'health', 'news', 'education', 'sports' & 0.281, 0.103, 0.098, 0.084 \\
today director members meeting board & 0.034 & 2.169 & 0.624 & 'business', 'news', 'education', 'government' & 0.312, 0.155, 0.131, 0.094 \\
fucking shit fuck i'm don't & 0.033 & 1.965 & 0.869 & 'comedy', 'society', 'news', 'leisure' & 0.447, 0.16, 0.068, 0.066 \\
people business i'm coach you're & 0.032 & 1.417 & 0.885 & 'business', 'education', 'health', 'society' & 0.563, 0.203, 0.096, 0.044 \\
open people back covid work & 0.032 & 2.023 & 0.571 & 'news', 'business', 'society', 'education' & 0.352, 0.247, 0.097, 0.047 \\
friends friend girl don't i'm & 0.032 & 2.042 & 0.517 & 'society', 'comedy', 'education', 'health' & 0.4, 0.157, 0.113, 0.057 \\
business businesses small company clients & 0.031 & 0.939 & 1.290 & 'business', 'education', 'news', 'society' & 0.788, 0.068, 0.042, 0.033 \\
wear wearing red shoes blue & 0.031 & 2.451 & 0.314 & 'arts', 'society', 'leisure', 'comedy' & 0.216, 0.139, 0.104, 0.101 \\
community people support group work & 0.031 & 2.289 & 0.273 & 'business', 'society', 'news', 'education' & 0.243, 0.174, 0.15, 0.098 \\
back looked eyes harry asked & 0.030 & 2.020 & 1.263 & 'arts', 'fiction', 'kids', 'education' & 0.36, 0.184, 0.112, 0.087 \\
city building place town street & 0.030 & 2.434 & 0.391 & 'society', 'business', 'news', 'education' & 0.257, 0.109, 0.105, 0.092 \\
company companies technology business industry & 0.029 & 1.354 & 1.223 & 'business', 'technology', 'news', 'education' & 0.631, 0.148, 0.068, 0.04 \\
car drive driving cars road & 0.029 & 2.294 & 0.549 & 'leisure', 'society', 'business', 'comedy' & 0.319, 0.116, 0.105, 0.092 \\
writing write read wrote paper & 0.029 & 2.091 & 0.573 & 'arts', 'education', 'society', 'business' & 0.342, 0.175, 0.147, 0.075 \\
people mask masks wear wearing & 0.029 & 2.457 & 0.203 & 'news', 'society', 'comedy', 'health' & 0.193, 0.186, 0.1, 0.089 \\
coffee drink it's beer drinking & 0.028 & 2.459 & 0.366 & 'arts', 'comedy', 'society', 'leisure' & 0.178, 0.162, 0.155, 0.095 \\
movie movies it's watch film & 0.028 & 1.399 & 1.733 & 'tv', 'comedy', 'arts', 'society' & 0.667, 0.086, 0.048, 0.042 \\
show season episode watch watching & 0.027 & 1.730 & 1.338 & 'tv', 'society', 'comedy', 'leisure' & 0.55, 0.087, 0.086, 0.067 \\
covid virus coronavirus people cases & 0.027 & 1.983 & 0.785 & 'news', 'health', 'society', 'education' & 0.45, 0.116, 0.092, 0.071 \\
store buy amazon shop sell & 0.027 & 2.029 & 0.474 & 'business', 'society', 'sports', 'news' & 0.458, 0.079, 0.069, 0.068 \\
food eat restaurant chicken pizza & 0.027 & 2.384 & 0.354 & 'arts', 'society', 'comedy', 'business' & 0.226, 0.158, 0.148, 0.071 \\
human idea world philosophy theory & 0.027 & 2.178 & 0.372 & 'society', 'religion', 'education', 'arts' & 0.235, 0.198, 0.187, 0.088 \\
god we're life today it's & 0.026 & 0.388 & 1.438 & 'religion', 'education', 'society', 'business' & 0.926, 0.019, 0.018, 0.009 \\
leadership team work people leaders & 0.026 & 1.206 & 1.083 & 'business', 'education', 'society', 'technology' & 0.7, 0.116, 0.038, 0.028 \\
god faith prayer pray god's & 0.026 & 0.707 & 1.174 & 'religion', 'society', 'education', 'arts' & 0.844, 0.048, 0.044, 0.015 \\
research university science professor work & 0.025 & 2.255 & 0.764 & 'science', 'education', 'business', 'society' & 0.229, 0.198, 0.126, 0.113 \\
mental health anxiety stress people & 0.025 & 1.919 & 0.677 & 'health', 'education', 'society', 'religion' & 0.408, 0.167, 0.128, 0.075 \\
fear feel afraid anxiety confidence & 0.025 & 1.928 & 0.376 & 'education', 'religion', 'health', 'business' & 0.267, 0.259, 0.132, 0.124 \\
phone app computer apple technology & 0.025 & 2.174 & 0.996 & 'technology', 'news', 'business', 'education' & 0.346, 0.118, 0.108, 0.103 \\
art design creative work artist & 0.025 & 1.801 & 0.967 & 'arts', 'business', 'society', 'education' & 0.503, 0.122, 0.106, 0.082 \\
black lives matter people racism & 0.024 & 2.508 & 0.058 & 'society', 'religion', 'education', 'news' & 0.205, 0.126, 0.097, 0.092 \\
college students school university student & 0.024 & 1.993 & 0.482 & 'education', 'society', 'business', 'news' & 0.417, 0.138, 0.132, 0.074 \\
black white people color racism & 0.024 & 2.232 & 0.246 & 'society', 'education', 'news', 'religion' & 0.315, 0.133, 0.117, 0.111 \\
relationship relationships marriage married person & 0.024 & 1.856 & 0.488 & 'society', 'religion', 'education', 'health' & 0.361, 0.187, 0.182, 0.11 \\
hair clean skin beauty cut & 0.024 & 2.412 & 0.292 & 'arts', 'society', 'education', 'comedy' & 0.183, 0.179, 0.127, 0.11 \\
government political people power state & 0.024 & 1.973 & 0.787 & 'news', 'society', 'religion', 'education' & 0.433, 0.133, 0.091, 0.085 \\
canada africa south country canadian & 0.024 & 2.288 & 0.389 & 'society', 'news', 'business', 'education' & 0.253, 0.225, 0.096, 0.088 \\
birthday ice party cream happy & 0.023 & 2.456 & 0.434 & 'society', 'kids', 'comedy', 'education' & 0.171, 0.158, 0.153, 0.085 \\
ghost dark find back monster & 0.023 & 2.294 & 1.047 & 'leisure', 'arts', 'fiction', 'comedy' & 0.273, 0.13, 0.124, 0.111 \\
history century world time years & 0.023 & 2.166 & 1.198 & 'history', 'society', 'education', 'arts' & 0.327, 0.149, 0.127, 0.099 \\
dog dogs animals cat animal & 0.023 & 2.355 & 0.867 & 'kids', 'education', 'society', 'science' & 0.287, 0.121, 0.113, 0.087 \\
crime case murder prison police & 0.023 & 2.320 & 1.219 & 'true crime', 'society', 'news', 'history' & 0.22, 0.189, 0.18, 0.076 \\
god lord jesus pray father & 0.023 & 0.574 & 1.276 & 'religion', 'society', 'education', 'business' & 0.871, 0.048, 0.024, 0.016 \\
day father mom dad mother & 0.023 & 2.032 & 0.346 & 'religion', 'society', 'education', 'kids' & 0.378, 0.17, 0.107, 0.09 \\
jack i'll music charlie sir & 0.023 & 1.822 & 1.890 & 'fiction', 'arts', 'comedy', 'society' & 0.431, 0.199, 0.112, 0.057 \\
energy spiritual life soul body & 0.023 & 1.554 & 0.606 & 'religion', 'education', 'health', 'society' & 0.486, 0.19, 0.136, 0.087 \\
school students learning education teacher & 0.023 & 1.466 & 1.035 & 'education', 'news', 'society', 'business' & 0.644, 0.066, 0.065, 0.057 \\
social people culture society change & 0.023 & 2.257 & 0.432 & 'society', 'education', 'business', 'news' & 0.211, 0.168, 0.151, 0.134 \\
water fish boat sea fishing & 0.022 & 2.511 & 0.329 & 'sports', 'society', 'religion', 'science' & 0.243, 0.111, 0.086, 0.083 \\
money financial credit bank tax & 0.022 & 1.384 & 0.929 & 'business', 'education', 'news', 'society' & 0.635, 0.141, 0.06, 0.048 \\
eat bread make milk food & 0.022 & 2.428 & 0.361 & 'arts', 'society', 'comedy', 'health' & 0.218, 0.123, 0.123, 0.102 \\
women men sex woman sexual & 0.022 & 2.082 & 0.459 & 'society', 'health', 'education', 'religion' & 0.319, 0.214, 0.133, 0.067 \\
joe laugh show nick comedy & 0.022 & 1.933 & 0.890 & 'comedy', 'society', 'tv', 'arts' & 0.487, 0.1, 0.08, 0.053 \\
china countries chinese world country & 0.021 & 1.927 & 0.761 & 'news', 'society', 'education', 'business' & 0.422, 0.164, 0.113, 0.079 \\
song music songs album listen & 0.021 & 1.971 & 1.157 & 'music', 'society', 'news', 'comedy' & 0.433, 0.15, 0.074, 0.073 \\
laughing yeah laughter laughs i'm & 0.020 & 2.408 & 0.503 & 'comedy', 'society', 'tv', 'leisure' & 0.259, 0.187, 0.085, 0.074 \\
war military army battle force & 0.020 & 2.499 & 0.650 & 'history', 'society', 'news', 'education' & 0.187, 0.144, 0.112, 0.11 \\
music band play rock album & 0.020 & 1.556 & 1.817 & 'music', 'arts', 'society', 'comedy' & 0.608, 0.083, 0.081, 0.039 \\
word bible chapter verse god & 0.020 & 0.549 & 1.303 & 'religion', 'society', 'education', None & 0.877, 0.05, 0.027, 0.009 \\
word english words language speak & 0.020 & 1.644 & 0.842 & 'education', 'society', 'arts', 'business' & 0.579, 0.117, 0.054, 0.047 \\
game games play playing played & 0.020 & 1.501 & 1.744 & 'leisure', 'tv', 'comedy', 'news' & 0.653, 0.054, 0.053, 0.037 \\
law legal insurance lawyer attorney & 0.020 & 1.818 & 0.704 & 'business', 'education', 'news', 'society' & 0.453, 0.155, 0.145, 0.068 \\
sports team sport play playing & 0.020 & 1.490 & 1.031 & 'sports', 'health', 'education', 'business' & 0.625, 0.077, 0.069, 0.052 \\
travel trip flight plane hotel & 0.019 & 2.113 & 0.465 & 'society', 'business', 'leisure', 'news' & 0.378, 0.131, 0.122, 0.081 \\
people george black floyd police & 0.019 & 2.303 & 0.218 & 'society', 'news', 'religion', 'comedy' & 0.258, 0.176, 0.114, 0.083 \\
god jesus christ sin grace & 0.019 & 0.405 & 1.422 & 'religion', 'society', 'education', 'arts' & 0.92, 0.02, 0.016, 0.012 \\
marketing brand content business digital & 0.019 & 1.082 & 1.209 & 'business', 'education', 'arts', 'society' & 0.758, 0.05, 0.037, 0.037 \\
trump president vote donald election & 0.019 & 1.320 & 1.391 & 'news', 'society', 'comedy', 'religion' & 0.693, 0.075, 0.039, 0.037 \\
american history states united america & 0.018 & 2.315 & 0.551 & 'news', 'society', 'history', 'education' & 0.206, 0.175, 0.137, 0.117 \\
church we're morning worship pray & 0.018 & 0.257 & 1.547 & 'religion', 'society', 'education', 'arts' & 0.952, 0.011, 0.007, 0.007 \\
film films movie director hollywood & 0.018 & 1.594 & 1.478 & 'tv', 'arts', 'society', 'news' & 0.589, 0.112, 0.066, 0.055 \\
energy air water power gas & 0.018 & 2.258 & 0.574 & 'business', 'education', 'news', 'technology' & 0.264, 0.167, 0.115, 0.093 \\
training fitness gym yoga exercise & 0.018 & 1.416 & 1.244 & 'health', 'education', 'sports', 'business' & 0.623, 0.098, 0.092, 0.063 \\
state county governor news public & 0.017 & 1.329 & 1.457 & 'news', 'government', 'society', 'business' & 0.673, 0.071, 0.055, 0.047 \\
church pastor ministry people churches & 0.017 & 0.636 & 1.230 & 'religion', 'society', 'education', 'business' & 0.871, 0.034, 0.023, 0.021 \\
i'm y'all shit ain't don't & 0.017 & 2.107 & 0.730 & 'society', 'music', 'comedy', 'news' & 0.269, 0.201, 0.161, 0.089 \\
farm plant plants garden food & 0.017 & 2.385 & 0.476 & 'leisure', 'education', 'news', 'business' & 0.171, 0.152, 0.122, 0.122 \\
running run bike race miles & 0.017 & 1.776 & 0.779 & 'sports', 'health', 'society', 'education' & 0.464, 0.21, 0.075, 0.053 \\
health care healthcare services covid & 0.016 & 2.060 & 0.696 & 'health', 'news', 'business', 'education' & 0.285, 0.18, 0.167, 0.101 \\
australia bit australian it's zealand & 0.016 & 2.375 & 0.275 & 'sports', 'society', 'news', 'comedy' & 0.191, 0.167, 0.165, 0.095 \\
jesus john disciples chapter matthew & 0.016 & 0.305 & 1.511 & 'religion', 'education', 'society', 'arts' & 0.943, 0.012, 0.01, 0.01 \\
weight food body eating eat & 0.016 & 1.459 & 1.231 & 'health', 'education', 'society', 'business' & 0.627, 0.122, 0.063, 0.042 \\
market economy markets year stock & 0.016 & 0.981 & 1.335 & 'business', 'news', 'education', 'society' & 0.713, 0.18, 0.024, 0.015 \\
body breath feel meditation breathe & 0.016 & 1.671 & 0.780 & 'health', 'religion', 'education', 'society' & 0.401, 0.238, 0.175, 0.063 \\
police officers officer law protests & 0.015 & 1.750 & 0.918 & 'news', 'society', 'comedy', 'education' & 0.536, 0.151, 0.05, 0.043 \\
coach football college state year & 0.015 & 0.825 & 1.652 & 'sports', 'news', 'society', 'education' & 0.812, 0.076, 0.026, 0.023 \\
space earth planet universe science & 0.015 & 2.552 & 0.723 & 'science', 'education', 'society', 'news' & 0.208, 0.127, 0.123, 0.081 \\
paul christ jesus chapter god & 0.015 & 0.240 & 1.561 & 'religion', 'education', 'society', None & 0.953, 0.012, 0.009, 0.008 \\
dance theater stage singing music & 0.015 & 1.917 & 1.047 & 'arts', 'music', 'society', 'education' & 0.469, 0.124, 0.089, 0.058 \\
lord god psalm praise psalms & 0.015 & 0.442 & 1.400 & 'religion', 'education', 'society', 'arts' & 0.91, 0.028, 0.022, 0.018 \\
sales marketing product customer website & 0.015 & 0.754 & 1.502 & 'business', 'education', 'technology', 'society' & 0.84, 0.049, 0.038, 0.019 \\
hospital doctor patients medical care & 0.014 & 2.252 & 0.466 & 'health', 'news', 'society', 'education' & 0.285, 0.15, 0.141, 0.104 \\
christian church faith religion religious & 0.014 & 1.017 & 0.973 & 'religion', 'society', 'education', 'news' & 0.763, 0.07, 0.055, 0.022 \\
michael jordan documentary dance sports & 0.014 & 1.605 & 0.959 & 'sports', 'news', 'society', 'tv' & 0.59, 0.096, 0.069, 0.051 \\
practice medicine medical health therapy & 0.014 & 1.674 & 1.057 & 'health', 'business', 'education', 'science' & 0.494, 0.162, 0.134, 0.056 \\
cancer pain blood disease surgery & 0.014 & 1.694 & 1.138 & 'health', 'education', 'science', 'society' & 0.548, 0.13, 0.079, 0.064 \\
real estate property market home & 0.014 & 0.841 & 1.372 & 'business', 'education', 'news', 'society' & 0.796, 0.081, 0.051, 0.031 \\
horse camp gun hunting bear & 0.013 & 1.857 & 0.776 & 'sports', 'education', 'society', 'leisure' & 0.523, 0.094, 0.075, 0.066 \\
data software cloud code machine & 0.013 & 1.537 & 1.913 & 'technology', 'business', 'education', 'news' & 0.54, 0.162, 0.092, 0.072 \\
disney florida park beach world & 0.013 & 1.716 & 0.870 & 'society', 'leisure', 'tv', 'news' & 0.569, 0.074, 0.065, 0.051 \\
nfl year team season game & 0.013 & 0.776 & 1.729 & 'sports', 'news', 'society', 'leisure' & 0.807, 0.112, 0.014, 0.014 \\
god genesis abraham adam earth & 0.012 & 0.558 & 1.315 & 'religion', 'society', 'education', 'arts' & 0.887, 0.034, 0.026, 0.014 \\
comic comics batman marvel max & 0.012 & 1.885 & 1.204 & 'tv', 'arts', 'leisure', 'news' & 0.393, 0.196, 0.138, 0.066 \\
game play playing cards card & 0.012 & 1.103 & 2.266 & 'leisure', 'fiction', 'comedy', 'games' & 0.759, 0.049, 0.045, 0.04 \\
sports teams season players hockey & 0.012 & 0.695 & 1.810 & 'sports', 'news', 'comedy', 'society' & 0.846, 0.077, 0.014, 0.013 \\
football game league back players & 0.011 & 0.706 & 1.799 & 'sports', 'news', 'leisure', 'comedy' & 0.854, 0.049, 0.019, 0.017 \\
lord david god moses israel & 0.011 & 0.372 & 1.459 & 'religion', 'education', 'society', 'arts' & 0.93, 0.017, 0.014, 0.01 \\
wedding karen jane call anna & 0.011 & 1.913 & 0.724 & 'news', 'society', 'leisure', 'health' & 0.422, 0.112, 0.098, 0.097 \\
brain system cells body cell & 0.011 & 1.985 & 0.978 & 'health', 'education', 'science', 'society' & 0.282, 0.25, 0.193, 0.056 \\
star wars episode anime trek & 0.011 & 1.719 & 1.483 & 'tv', 'leisure', 'comedy', 'fiction' & 0.514, 0.167, 0.06, 0.052 \\
nba team he's game basketball & 0.011 & 1.037 & 1.491 & 'sports', 'news', 'society', 'leisure' & 0.766, 0.085, 0.032, 0.025 \\
fight he's fighting fights boxing & 0.010 & 1.388 & 1.156 & 'sports', 'news', 'health', 'society' & 0.67, 0.076, 0.049, 0.042 \\
baseball players game league season & 0.010 & 0.830 & 1.678 & 'sports', 'news', 'society', 'leisure' & 0.815, 0.083, 0.016, 0.015 \\
baby birth pregnant women pregnancy & 0.010 & 1.905 & 1.207 & 'kids', 'health', 'society', 'education' & 0.331, 0.285, 0.104, 0.078 \\
spirit holy jesus god pentecost & 0.010 & 0.272 & 1.533 & 'religion', 'society', 'education', 'arts' & 0.948, 0.011, 0.01, 0.007 \\
health body vitamin skin system & 0.010 & 1.473 & 1.275 & 'health', 'education', 'business', 'arts' & 0.629, 0.104, 0.055, 0.041 \\
court india case supreme indian & 0.010 & 1.590 & 1.199 & 'news', 'education', 'society', 'government' & 0.602, 0.07, 0.06, 0.059 \\
draft he's year pick round & 0.010 & 0.556 & 1.950 & 'sports', 'news', 'leisure', 'society' & 0.87, 0.079, 0.016, 0.008 \\
golf play tennis ball playing & 0.010 & 0.857 & 1.654 & 'sports', 'news', 'comedy', 'society' & 0.828, 0.036, 0.024, 0.023 \\
god chapter people lord verse & 0.009 & 0.370 & 1.457 & 'religion', 'society', 'education', 'news' & 0.932, 0.018, 0.014, 0.009 \\
king queen prince kings kingdom & 0.009 & 1.596 & 0.758 & 'religion', 'arts', 'education', 'kids' & 0.584, 0.08, 0.078, 0.074 \\
wrestling match show wwe ring & 0.009 & 0.985 & 1.582 & 'sports', 'tv', 'news', 'comedy' & 0.785, 0.039, 0.037, 0.037 \\
peter jesus acts paul chapter & 0.009 & 0.293 & 1.528 & 'religion', 'arts', 'education', 'society' & 0.944, 0.014, 0.013, 0.01 \\
security information data cyber privacy & 0.009 & 1.741 & 1.490 & 'technology', 'news', 'business', 'education' & 0.395, 0.241, 0.18, 0.048 \\
wine cannabis smoking smoke cbd & 0.008 & 2.300 & 0.427 & 'arts', 'society', 'business', 'leisure' & 0.259, 0.17, 0.094, 0.092 \\
thy mary father thou holy & 0.008 & 0.666 & 1.274 & 'religion', 'education', 'kids', 'arts' & 0.854, 0.044, 0.037, 0.024 \\
kroger save month mint savings & 0.008 & 2.545 & 0.278 & 'comedy', 'society', 'sports', 'news' & 0.174, 0.148, 0.126, 0.107 \\
jewish daniel israel jews people & 0.008 & 0.815 & 1.123 & 'religion', 'news', 'society', 'education' & 0.81, 0.05, 0.042, 0.035 \\
michigan vegas casino bet las & 0.008 & 2.513 & 0.476 & 'society', 'sports', 'news', 'fiction' & 0.158, 0.149, 0.127, 0.119 \\
it's free that's code visit & 0.007 & 2.122 & 0.590 & 'sports', 'news', 'comedy', 'society' & 0.378, 0.15, 0.104, 0.103 \\
bitcoin gold oil trading market & 0.007 & 1.289 & 1.257 & 'business', 'news', 'technology', 'education' & 0.581, 0.17, 0.138, 0.047 \\
moon sign card energy astrology & 0.007 & 1.675 & 0.529 & 'religion', 'education', 'society', 'health' & 0.512, 0.141, 0.14, 0.051 \\
race racing nascar track races & 0.006 & 1.244 & 1.365 & 'sports', 'news', 'leisure', 'society' & 0.527, 0.307, 0.099, 0.015 \\
patients study drug clinical risk & 0.006 & 1.455 & 1.666 & 'health', 'science', 'education', 'news' & 0.516, 0.227, 0.1, 0.046 \\
krishna hai para con una & 0.006 & 2.109 & 0.295 & 'religion', 'education', 'society', 'arts' & 0.32, 0.184, 0.157, 0.065 \\
sports app that's you're podcast & 0.005 & 1.794 & 0.863 & 'sports', 'news', 'comedy', 'society' & 0.476, 0.147, 0.123, 0.089 \\
language speaking foreign spanish speaks & 0.004 & 2.244 & 0.167 & 'religion', 'education', 'society', 'news' & 0.252, 0.203, 0.142, 0.076 \\
allah muslim islam ramadan prophet & 0.003 & 1.044 & 1.037 & 'religion', 'education', 'society', 'kids' & 0.522, 0.4, 0.034, 0.008 \\
yeah leroy laughter i'm craft & 0.003 & 0.715 & 3.351 & 'technology', 'fiction', 'comedy', 'education' & 0.838, 0.083, 0.028, 0.009 \\
it's that's person torah hashem & 0.003 & 0.594 & 1.292 & 'religion', 'education', 'arts', 'society' & 0.815, 0.151, 0.015, 0.005 \\
\caption{Top words, categories, and category distribution entropy for each topic, sorted by average category proportion across all podcasts.} 
\label{tab:allstorms}
\end{longtable}
\end{tiny}

\twocolumn

\end{document}